\documentclass[letterpaper, 10 pt, journal, twoside]{IEEEtran}

\IEEEoverridecommandlockouts                              

\usepackage{amsmath} 
\usepackage{amssymb}  

\usepackage[dvipsnames]{xcolor}
\usepackage{soul}
\usepackage{graphicx}
\usepackage{url}

\usepackage[inline]{enumitem}
\usepackage{algorithm}
\usepackage{algpseudocode}

\usepackage{tabularray}
\usepackage{booktabs}

\usepackage{tikz}
\usetikzlibrary{bayesnet}
\usetikzlibrary{arrows.meta, positioning}

\usepackage{cite}

\makeatletter
\AddToHook{env/algorithmic/begin}{\def\@currentcounter{ALG@line}}
\makeatother

\usepackage{cleveref}

\usepackage{aliascnt}

\crefformat{section}{\S#2#1#3} 
\crefformat{subsection}{\S#2#1#3}
\crefformat{subsubsection}{\S#2#1#3}

\crefname{algorithm}{alg.}{algs.}
\Crefname{algorithm}{Algorithm}{Algorithms}
\crefname{ALG@line}{line}{lines}
\Crefname{ALG@line}{Line}{Lines}

\usepackage{siunitx}

\algnewcommand\algorithmicforeach{\textbf{for each}}
\algdef{S}[FOR]{ForEach}[1]{\algorithmicforeach\ #1\ \algorithmicdo}

\usepackage[normalem]{ulem}

\newcommand{\dashulinecolor}[3]{%
  \tikz[baseline=(text.base)]{
    \node[inner sep=0pt, outer sep=0pt] (text) {\textcolor{#2}{#1}};
    \draw[#3, dashed, thick] 
      ([yshift=-2pt]text.south west) -- 
      ([yshift=-2pt]text.south east);
  }%
}

\usepackage{nth}

\usepackage{duckuments}

\definecolor{black}{rgb}{0,0,0}

\definecolor{plotblue}{HTML}{DAF3FD}
\definecolor{plotred}{HTML}{FFD8D8}
\definecolor{plotpurple}{HTML}{D7D7FD}
\definecolor{plotviolet}{HTML}{E9D8FC}
\definecolor{plotgreen}{HTML}{D7FCE8}

\definecolor{darkred1}{HTML}{A52714}
\definecolor{darkgreen1}{HTML}{0B8043}
\definecolor{darkblue1}{HTML}{174EA6}

\newcommand{\colul}[2]{\setulcolor{#1}\ul{#2}}

\newcommand{\plotrefblue}[1]{\colul{plotblue}{#1}}
\newcommand{\plotrefred}[1]{\colul{plotred}{#1}}
\newcommand{\plotrefpurple}[1]{\colul{plotpurple}{#1}}
\newcommand{\plotrefviolet}[1]{\colul{plotviolet}{#1}}
\newcommand{\plotrefgreen}[1]{\colul{plotgreen}{#1}}

\begin{document}

\title{\LARGE \bf
Posterior-driven Heuristic Support Adaptation in a Probabilistic Treatment of Real2Sim2Real for Vision-Driven Deformable Linear Object Manipulation
}

\author{Georgios Kamaras$^{1,2}$, Craig Innes$^{1}$, and Subramanian Ramamoorthy$^{1,3}$
\thanks{$^{1}$School of Informatics, The University of Edinburgh, EH8 9AB, UK.\\{\tt\small \{gkamaras, cinnes, s.ramamoorthy\}@ed.ac.uk}%
}
\thanks{$^{2}$Corresponding author. Work supported by the Engineering and Physical Sciences Research Council (EPSRC), as part of the CDT in RAS at Heriot-Watt University and The University of Edinburgh.%
}
\thanks{$^{3}$Work supported by a UKRI Turing AI World Leading Researcher Fellowship on AI for Person-Centred and Teachable Autonomy (grant EP/Z534833/1).%
}
\thanks{For the purpose of open access, the authors have applied a Creative Commons Attribution (CC BY) licence to any Author Accepted Manuscript version arising from this submission.}
}

\maketitle

\begin{abstract}

Likelihood-free inference (LFI) enables system identification in complex tasks via black-box modelling, abstracting nonlinearity and stochasticity, and infers a domain distribution for adapting agents to parametric deployment conditions.
LFI assumes an arbitrary support for sampling, which remains fixed as the initial generic prior is refined to increasingly descriptive posteriors. 
Misspecified support can therefore yield suboptimal yet overconfident posteriors.
We address this issue by using the posterior of an inference step to guide the adaptation of the support using three illustrative heuristics: EDGE, MODE, and CENTRE. 
Each heuristic interprets the updated belief and enables support adaptation alongside posterior inference. 
For illustrative purposes, we first study misspecified support in LFI and evaluate the utility of our heuristics using stochastic dynamical benchmarks.
We then evaluate posterior-driven heuristic support adaptation for parameter inference and policy learning in a dynamic deformable linear object (DLO) manipulation task. 
Inference results in a finer length and stiffness classification for a parametric set of DLOs. When the resulting posteriors are used as domain distributions for sim-based policy learning, they lead to more robust object-centric agent performance.

\end{abstract}

\def\abstractname{Note to Practitioners}
\begin{abstract}

This paper was motivated by the difficulty of defining the parameter space, or \emph{domain support}, for the probabilistic inference of the length and stiffness of a deformable linear object through vision.
This definition is nontrivial when inferring the distribution that will calibrate a simulation for policy learning, as we may lack intrinsic knowledge of the real-world system and how real physical properties correspond to the simulation parameters. 
Current methods assume an arbitrary \emph{`best guess'} support, but this is susceptible to misspecification for systems of increased complexity and stochasticity. 
We study how the inferred posteriors may indicate the need to adapt the support to more accurately reflect the range of possible parameterisations. 
We define a set of heuristics to iteratively adapt the support in tandem with posterior inference.
Our experiments show that this increases the reliability of both parameter inference and agent skill adaptation. 
Our heuristics can be used as baselines for future methods that explore more formal approaches to support adaptation, or automatically extract the rule that should guide it. 
The potential impact of our methodology spans from closing the `reality gap' to robustly conditioning autonomous agents for an estimated context under unknown system dynamics.

\end{abstract}


\section{Introduction}
\label{sec:intro}

\begin{figure}[t]
    \centering
    \includegraphics[width=1.0\columnwidth]{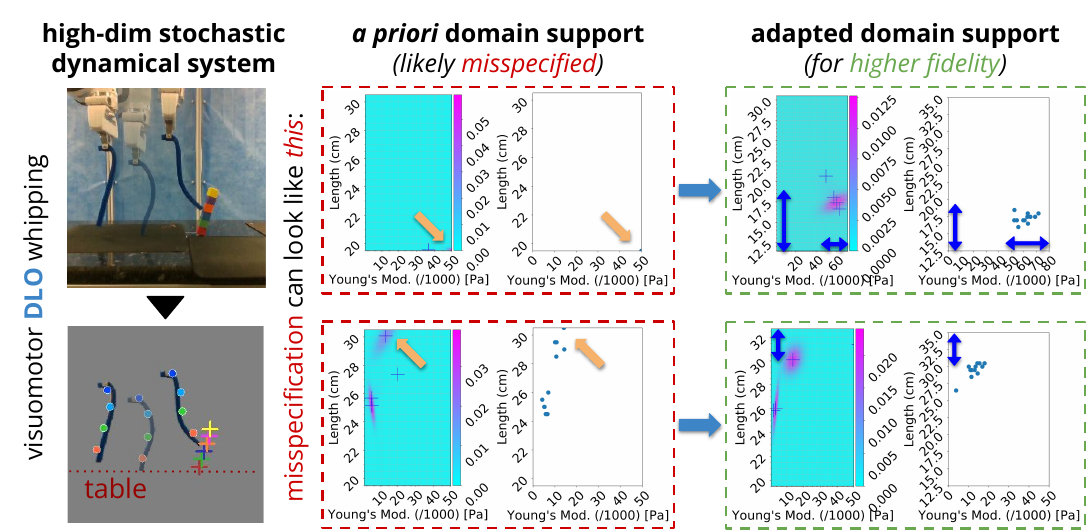}
    \caption{
    The support misspecification issue in a visuomotor DLO whipping task (\textcolor{darkblue1}{left}). 
    We see how LFI struggles to accurately infer the Young's modulus and length on a \emph{misspecified support}, and the impact on domain samples from the posterior (\textcolor{darkred1}{centre}). 
    We then see how \emph{support adaptation} leads to more accurate inference and more descriptive domain samples (\textcolor{darkgreen1}{right}). Orange arrows denote posterior mass accumulation on domain boundaries due to misspecification. Blue arrows denote the corresponding adaptations, which \emph{stretch} the domain.
    }
    \label{fig:motivating-fig}
\end{figure}

\IEEEPARstart{A}{ccurate} control of deformable linear objects (DLOs) is critical in assistive, manufacturing, and surgical robotics~\cite{arriola2020modeling, yin2021modeling, caporali2024robotic}. 
Even similarly shaped DLOs can behave very differently under manipulation due to variations in length and stiffness. 
We study a visuomotor DLO whipping task, in which a DLO drawn from such a parametric set is guided near a cube stack to dislodge (\emph{whip}) the top cube (\cref{fig:motivating-fig}, left), highlighting the sensitivity of task outcomes to object properties. 
To avoid the inefficiency of high-dimensional pixel inputs, we track the DLO and cubes using inferred keypoints, yielding compact visual state representations~\cite{li2020causal, antonova2022bayesian, zhang2024adaptigraph, kulkarni2019unsupervised}.

Modern simulators enable large-scale data collection and policy training for dynamic DLO manipulation~\cite{kim2022dynamic, chi2024iterative, haiderbhai2024sim2real, kamaras2025distributional}. 
Closing the \emph{reality gap}~\cite{aljalbout2025reality} requires calibrating simulator parameters $\boldsymbol{\theta}$ to the real system (\emph{Real2Sim})~\cite{mehta2021user, liang2020learning}. 
Likelihood-free inference (LFI)~\cite{papamakarios2016fast}, including BayesSim and related methods~\cite{ramos2019bayessim, heiden2022probabilistic, ren2023adaptsim, memmel2024asid, maddukuri2025sim}, models the multimodal posterior $\hat{p}(\boldsymbol{\theta})$ as a mixture of Gaussians, whose modes represent parameter \emph{hypotheses} and variances encode uncertainty.

This probabilistic framework integrates naturally with domain randomisation (DR)~\cite{tobin2017domain}, where RL policies are trained in simulators parameterised by samples from a \emph{domain distribution}~\cite{muratore2022robot} to improve real-world deployment (\emph{Sim2Real}) robustness~\cite{liang2024real, haiderbhai2024sim2real}. 
Wide, high-entropy distributions yield \emph{generalist} agents, while narrow, concentrated ones produce \emph{specialists}~\cite{tang2024automate}. 
Although generalists suit low-stakes, unstructured tasks, specialists are more efficient, predictable, and certifiable in high-stakes structured settings~\cite{haiderbhai2024sim2real}.

The integration of the above two methods constitutes a \emph{Real2Sim2Real} framework.
This end-to-end approach requires an \emph{a priori} sampling prior for LFI or DR, typically a uniform distribution with finite rectangular support $\Theta=[\boldsymbol{\theta}_{\min},\boldsymbol{\theta}_{\max}]$~\cite{ramos2019bayessim, matl2020inferring, antonova2022bayesian, heiden2022probabilistic, maddukuri2025sim}, which we denote as $p_{(\Theta)}$ in this paper. 
Defining this \emph{domain support} is often ad hoc and can lead to \emph{prior misspecification}~\cite{rainforth2024modern}. 
Moreover, in LFI the support $\Theta$ usually remains fixed as the prior is refined into concentrated MoGs, which is especially problematic for Real2Sim inference where system knowledge is limited.

To illustrate, in our DLO whipping task we aim to strike the top cube without disturbing the stack. 
The relevant physical parameters---whip length and stiffness---are unknown and difficult to bound: length may be underspecified, and stiffness remains uncertain even when the base material is known. 
Manufacturing can alter material properties~\cite{liao2020ecoflex}, and simulator \emph{realism} need not imply \emph{physical accuracy}~\cite{hofer2020perspectives}, as the parameters matching observed behaviour may differ from the object’s true properties.

Thus, defining an accurate support $\Theta$ for the LFI of $\boldsymbol{\theta}$ is difficult and risks excluding useful regions while biassing parameter inference, and consequently policy learning, through incorrect assumptions.
If the true parameters lie outside $\Theta$, posterior mass accumulates at its boundaries (\cref{fig:motivating-fig}, centre), yielding biased estimates as LFI assigns observations to the nearest feasible region.
When such posteriors are used for DR, this bias directly propagates to policy learning.

One may ask, \emph{``Why not use the widest support possible?''}. While this avoids excluding useful parameterisations, it often \emph{reduces data efficiency and quality} (\cref{subsec:widest-prior}). 
For instance, physics solvers impose feasibility limits (e.g. minimum stiffness), below which simulation becomes unstable due to mesh collapse~\cite{kim2022dynamic}, yielding invalid or uninformative data. 
We therefore restrict inference to a \emph{physically feasible} domain $\Phi \supset \Theta$.

These challenges motivate LFI extensions that iteratively adapt a misspecified support $\Theta$. 
Our key insight is that the shape and location of a parametric (MoG) posterior indicate whether, and in which direction, the support should be adjusted to better estimate $\boldsymbol{\theta}$ (\cref{fig:motivating-fig}). 
This approach is more efficient than formal \emph{information acquisition}~\cite{greenhill2020bayesian}, which is intractable here, since each candidate support would require a full, costly inference run.

In this paper, we make the following contributions:
\begin{enumerate}
    \item We expose how \textbf{support misspecification leads to suboptimal posterior inference} despite \emph{perceived certainty} of the results in 
    a realistic physical system with stochastic dynamics that can be tractably simulated. For illustrative purposes, 
    we use the Lotka-Volterra model~\cite{lotka1920analytical} (\cref{sec:problem-overview}).

    \item We show how \textbf{key posterior features}, such as probability mass accumulation and mode shift, \textbf{can guide the adaptation} of a misspecified support by proposing three BayesSim variants (EDGE, MODE, CENTRE), each using a different \textbf{support adaptation heuristic} (\cref{sec:heur-supp-opt-methods}, \cref{sec:lv-poc-experiment}).

    \item We use the most robust variant in a \textbf{Real2Sim2Real} framework for DLO whipping. We use the LFI posteriors, \textbf{with and without support adaptation}, for DR in RL policy learning in simulation and evaluate these policies in the real-world (\cref{sec:r2s2r-dlo-whip-exper}, \cref{sec:dlo-results}). 
    
\end{enumerate}

Our experiments show that support adaptation during LFI refines the inference of physical properties among a parametric set of similarly shaped (real) DLOs. Through DR, this fine inference translates into a stronger object-centric agent specialisation with measurable real-world performance impact.

\section{Background}
\label{sec:background}

\subsection{Likelihood-free inference}
\label{subsec:lfi-prelim}

LFI, being a simulation-based inference (SBI)~\cite{cranmer2020frontier} technique, 
treats a simulator as a black-box generative model $g$ with an intractable likelihood. Given parameters $\boldsymbol{\theta}$, the simulator produces an observation $\mathbf{x} = g(\boldsymbol{\theta})$, implicitly defining a likelihood $p(\mathbf{x} \mid \boldsymbol{\theta})$ that cannot be evaluated but can be sampled by running the simulator~\cite{papamakarios2016fast}. 

Given a real trajectory $\mathbf{x}^r$, we narrow the reality gap by inferring the parameters $\boldsymbol{\theta}$ most likely to reproduce it in simulation. 
This defines the inverse problem of approximating the posterior $\hat{p}(\boldsymbol{\theta} \mid \mathbf{x} = \mathbf{x}^r)$. 

Among various LFI approaches to this problem~\cite{papamakarios2016fast, greenberg2019icml, papamakarios2019sequential, durkan2020icml, papamakarios2021normalizing, BoeltsDeistler_sbi_2025}, BayesSim~\cite{ramos2019bayessim} is particularly prominent in robotics~\cite{barcelos2020disco, possas2020online, matl2020inferring, matl2020stressd, antonova2022bayesian} due to its algorithmic simplicity and practical parametric representation of the posterior as a mixture of Gaussians (MoG). MoGs provide tractable and interpretable representations of alternative hypotheses that can be directly sampled or mathematically manipulated, making them well suited to domain randomisation (\cref{subsec:dr-prelim}). Particle-based posterior approximation alternatives~\cite{heiden2021disect, heiden2022probabilistic} may be more descriptive but are less convenient as generative models for domain sampling. We therefore focus on BayesSim as our baseline due to its conceptual simplicity and demonstrated effectiveness for Real2Sim inference in robotics. 

BayesSim (\cref{alg:bsim}) approximates the $\hat{p}(\boldsymbol{\theta} \mid \mathbf{x} = \mathbf{x}^r)$ posterior by learning the conditional density function (CDF) $q_{\phi}(\boldsymbol{\theta} \mid \mathbf{x})$, parameterised by $\boldsymbol{\phi}$. 
The CDF is modelled as a MoG, learned with a mixture density neural network (MDNN)~\cite{bishop1994mixture}. Inputs $\mathbf{x}$ may be raw trajectories, summary statistics $\psi(\cdot)$, or kernel features.

BayesSim first requires training $q_{\phi}(\boldsymbol{\theta} \mid \mathbf{x})$ on a dataset of $N$ pairs $(\boldsymbol{\theta}_n, \mathbf{x}_n)$, where the parameters $\boldsymbol{\theta}_n$ are drawn from a \emph{proposal prior} $\Tilde{p}(\boldsymbol{\theta})$ and the observation trajectories $\mathbf{x}_n$ are generated by running $g(\boldsymbol{\theta}_n)$ for the duration of a simulated episode and collecting the respective state-action pairs.
Then, given a real-world trajectory $\mathbf{x}^r$, it estimates the posterior as:
\begin{equation}
    \hat{p}(\boldsymbol{\theta} \mid \mathbf{x} = \mathbf{x}^r) \propto \frac{p(\boldsymbol{\theta})}{\Tilde{p}(\boldsymbol{\theta})} q_{\phi}(\boldsymbol{\theta} \mid \mathbf{x} = \mathbf{x}^r) \text{,}
    \label{eq:bsim-posterior}
\end{equation}
where the desirable prior $p(\boldsymbol{\theta})$ can differ from the proposal prior. If $\Tilde{p}(\boldsymbol{\theta}) = p(\boldsymbol{\theta})$, then $ \hat{p}(\boldsymbol{\theta} \mid \mathbf{x} = \mathbf{x}^r) \propto q_{\phi}(\boldsymbol{\theta} \mid \mathbf{x} = \mathbf{x}^r)$.
\Cref{eq:bsim-posterior} also enables \emph{iterative} posterior refinement (\cref{alg:bsim}, \cref{alg:bsim:line:cdf-update} and \cref{fig:bayessim-sequential-loop}).  
This progressively adapts the sampling distribution $p(\boldsymbol{\theta})$ to regions with a higher posterior density~\cite{geweke2019sequentially}, which yield simulated observations $\mathbf{x}$ similar to $\mathbf{x}^r$~\cite{antonova2022bayesian, possas2020online}. 

\begin{figure}[t]
\centering
\resizebox{0.95\columnwidth}{!}{%
\begin{tikzpicture}[
    node distance=1.0cm and 1.8cm,
    every node/.style={font=\large}
  ]

  \node[latent] (proposal) {$\Tilde{p}(\boldsymbol{\theta})$};
  \node[obs, above=of proposal] (xr) {$\mathbf{x}^r$};

  \node[latent, right=of proposal] (p1) {$\hat{p}_1(\boldsymbol{\theta} \mid \mathbf{x}^r)$};
  \node[latent, right=of p1] (p2) {$\hat{p}_2(\boldsymbol{\theta} \mid \mathbf{x}^r)$};
  \node[latent, right=of p2] (p3) {$\hat{p}_3(\boldsymbol{\theta} \mid \mathbf{x}^r)$};

  \node[obs, below=of proposal, yshift=-0.425cm] (theta0) {$\boldsymbol{\theta}_0$};
  \node[obs, below=of p1] (theta1) {$\boldsymbol{\theta}_1$};
  \node[obs, below=of p2] (theta2) {$\boldsymbol{\theta}_2$};
  \node[obs, below=of p3] (theta3) {$\boldsymbol{\theta}_3$};

  \node[obs, below=of theta0] (x0) {$\mathbf{x}_0$};
  \node[obs, below=of theta1] (x1) {$\mathbf{x}_1$};
  \node[obs, below=of theta2] (x2) {$\mathbf{x}_2$};
  \node[obs, below=of theta3] (x3) {$\mathbf{x}_3$};

  \draw[->,teal!70!black] (proposal) -- (theta0) node[midway,right] {\small sample};
  \draw[->,teal!70!black] (p1) -- (theta1) node[midway,right] {\small sample};
  \draw[->,teal!70!black] (p2) -- (theta2) node[midway,right] {\small sample};
  \draw[->,teal!70!black] (p3) -- (theta3) node[midway,right] {\small sample};

  \draw[->,bend right=15,blue!80!black] (x0) to node[midway,above,xshift=-0.35cm] {\small update} (p1);
  \draw[->,bend right=15,blue!80!black] (x1) to node[midway,above,xshift=-0.35cm] {\small update} (p2);
  \draw[->,bend right=15,blue!80!black] (x2) to node[midway,above,xshift=-0.35cm] {\small update} (p3);

  \draw[->,orange!80!black] (theta0) -- (x0) node[midway,right, 
      fill=white,
      fill opacity=0.7,
      text opacity=1,
      inner ysep=1pt
  ] {\small simulate};
  \draw[->,orange!80!black] (theta1) -- (x1) node[midway,right, 
      fill=white,
      fill opacity=0.7,
      text opacity=1,
      inner ysep=1pt
  ] {\small simulate};
  \draw[->,orange!80!black] (theta2) -- (x2) node[midway,right, 
      fill=white,
      fill opacity=0.7,
      text opacity=1,
      inner ysep=1pt
  ] {\small simulate};
  \draw[->,orange!80!black] (theta3) -- (x3) node[midway,right, 
      fill=white,
      fill opacity=0.7,
      text opacity=1,
      inner ysep=1pt
  ] {\small simulate};

  \draw[->,purple!70!black] (x0) -- (x1) node[midway,below] {\small \shortstack{augment\\$\mathcal{D} \gets \mathbf{x}_0$}};
  \draw[->,purple!70!black] (x1) -- (x2) node[midway,below] {\small \shortstack{augment\\$\mathcal{D} \gets \mathcal{D} \cup \mathbf{x}_1$}};
  \draw[->,purple!70!black] (x2) -- (x3) node[midway,below] {\small \shortstack{augment\\$\mathcal{D} \gets \mathcal{D} \cup \mathbf{x}_2$}};

  \draw[dotted,->] (xr) to [bend left=10] node[midway,below,text=gray!70!black] {\small compare} (p1);
  \draw[dotted,->] (xr) to [bend left=15] node[midway,below,text=gray!70!black] {\small compare} (p2);
  \draw[dotted,->] (xr) to [bend left=20] node[midway,below,text=gray!70!black] {\small compare} (p3);

  \node[const, right=of p3, xshift=-0.75cm] (dotsR) {$\ldots$};

\end{tikzpicture}
}
\caption{
Iterative (or \emph{sequential}) refinement of Bayesian posterior approximations. Each posterior $\hat{p}_t$ is used to sample new parameters, generate simulations, and retrain the density estimator. This adaptive loop densifies coverage in high-likelihood regions.
}
\label{fig:bayessim-sequential-loop}
\end{figure}

\begin{algorithm}[t]
\caption{Iterative Bayesian LFI for Real2Sim2Real}
\label{alg:bsim}
\begin{algorithmic}[1]
    \State \textbf{Given:} $\Tilde{p}(\boldsymbol{\theta}) \approx \mathit{U}$: Proposal prior; $N_{\text{LFI}}$: iterations; \\
           $\quad\quad\quad$ $\Theta_0 = [\boldsymbol{\theta}_{\min}, \boldsymbol{\theta}_{\max}]$: initial support bounds; \\ 
           $\quad\quad\quad$ $\Phi$: phys. feasible domain limits
    \\
    \State Assign reference prior $p_0 \gets \Tilde{p}$
    \State Train initial policy $\pi_{\boldsymbol{\beta}_0}(\mathbf{a}_t \mid \mathbf{s}_t)$, $\boldsymbol{\theta} \sim p_0$
    \State Run $1$ $\pi_{\boldsymbol{\beta}_0}$ rollout in the real env to collect $\mathbf{x}^r$
    \\
    \State \textbf{// 1. Real2Sim parameter inference (LFI)}
    \State $i \gets 0;\;\mathcal{D} \gets \{\};\;D \gets \lvert \boldsymbol{\theta}_{\min} \rvert$
    \While{$i < N_{\text{LFI}}$} \label{alg:bsim:line:lfi-iter}
        \State \textcolor{orange!70!black}{$\{(\boldsymbol{\theta}, \mathbf{x}^s)\}^N_i \gets$ Simulate $N$ $\pi_{\boldsymbol{\beta}_0}$ rollouts}, \textcolor{teal!70!black}{$\boldsymbol{\theta} \sim p_i$}
        \State \textcolor{purple!70!black}{$\mathcal{D} \gets \mathcal{D} \cup \{(\boldsymbol{\theta}, \mathbf{x}^s)\}^N_i$ \label{alg:bsim:line:dataset-update}
        \Comment{Dataset augmentation}}
        \State Train $q_{\phi}$ over $\mathcal{D}$
        \State \textcolor{gray!70!black}{$\hat{p}(\boldsymbol{\theta} \mid \mathbf{x} = \mathbf{x}^r) \propto (p_i(\boldsymbol{\theta}) \mathbin{/} \Tilde{p}(\boldsymbol{\theta})) q_{\phi}(\boldsymbol{\theta} \mid \mathbf{x} = \mathbf{x}^r)$} \label{alg:bsim:line:cdf-update}
        \If{\textcolor{cyan!60!black}{\texttt{Heuristic Support Adaptation}}}
            \State $\boldsymbol{\theta}_{\min}, \boldsymbol{\theta}_{\max} = \{\textcolor{cyan!60!black}{\text{EDGE, MODE, CENTRE}}\}(...)$\label{alg:bsim:hda-extension}
        \EndIf
        \State $i \gets i + 1$
        \State $\Theta_i \gets [\boldsymbol{\theta}_{\min}, \boldsymbol{\theta}_{\max}]$
        \Comment{Update bounds}
        \State \textcolor{blue!60!black}{$p_i \gets \hat{p}_{(\Theta_i)}$ \label{alg:bsim:line:p-update}
        \Comment{Update reference prior}}
    \EndWhile
    \\
    \State \textbf{// 2. Policy training in sim}
    \State Train policy $\pi_{\boldsymbol{\beta}_1}(\mathbf{a}_t \mid \mathbf{s}_t)$, $\boldsymbol{\theta} \sim \hat{p}$ \label{alg:bsim:line:posterior-policy-train}
    \\
    \State \textbf{// 3. Sim2Real policy deployment}
    \State Evaluate $\pi_{\boldsymbol{\beta}_1}$ in the real env by running $1$ $\pi_{\boldsymbol{\beta}_1}$ rollout
\end{algorithmic}
\end{algorithm}

\subsection{Adaptive domain randomisation}
\label{subsec:dr-prelim}

DR injects sufficient simulated variability during training so models generalise to real-world data~\cite{tobin2017domain}.
It does this by sampling simulator parameters $\boldsymbol{\theta}$ from a domain distribution.
However, combining the inherent instability of RL training with a wide uniform prior in DR can limit robustness~\cite{peng2018sim, possas2020online}. 
An LFI posterior offers a narrower and more accurate distribution, concentrated around the system's true parameters.

To our knowledge, we are the first to adapt the support of the parameter space for Real2Sim inference. Instead, prior \emph{adaptive DR} work trains \emph{generalist} agents robust to environment perturbations~\cite{akkaya2019solving, mehta2020active, josifovski2024continual}. These methods begin with an estimated domain range and typically \emph{expand} it during training~\cite{handa2023dextreme}, steering DR toward parameterisations that challenge the current model or policy. 
Their premise is that very wide domain distributions let the agent improve its behaviour within a rollout~\cite{akkaya2019solving}. However, this is infeasible for dynamic tasks like DLO whipping~\cite{zhang2021robots, lim2022real2sim2real, chi2024iterative}, which are too short (e.g., $<10\,\si{\sec}$ and far $<32$ steps) for such online adaptation.
Decoupling deployment conditions from policy learning also reduces the verifiability of behaviours in high-stakes tasks, such as medical automation~\cite{haiderbhai2024sim2real}.

We begin with a best estimate of the support~\cite{handa2023dextreme} and expand it~\cite{akkaya2019solving} when a heuristic condition is met. 
Unlike prior work, we perform LFI in an integrated Real2Sim2Real framework. This \emph{efficiently} and \emph{interpretably} yields a \emph{specialist} agent for a specific real environment~\cite{lim2022real2sim2real, memmel2024asid, zhang2024adaptigraph}, since we use inferred posteriors as the domain distribution during training. 
This aligns DR with our estimate of the deployment parameters and induces object-centric behaviour~\cite{kamaras2025distributional}.

\section{A demo of the support misspecification issue}
\label{sec:problem-overview}

We inspect the impact of a misspecified support by asking: 
\begin{enumerate}
    \item What does the misspecified support issue \textbf{look like}?
    \item Why not use the \textbf{widest support} possible?
\end{enumerate}
We focus our experiments to the Lotka-Volterra stochastic dynamical benchmark, as implemented in~\cite{papamakarios2016fast}, but a complementary study with the M/G/1 model is available in~\cref{appendix:mg1-case-study}.

\subsection{Lotka-Volterra predator-prey population model}
\label{subsec:lv-exper}

The Lotka-Volterra model is a lower-dimensional stochastic system describing predator ($X$)-prey ($Y$) interactions~\cite{lotka1920analytical}. It includes four event types---predator birth, predator death, prey birth, and prey death by predation---whose respective rates are the positive parameters $\boldsymbol{\theta} = \{ \theta_1, \theta_2, \theta_3, \theta_4 \}$.
Nature-like population circles arise only for specific $\boldsymbol{\theta}$, and many settings yield infeasible simulations, analogous to \emph{infeasible physical configurations} in robotic simulators. We therefore enforce a strict \emph{effective} support by clipping outlier domain samples. 

\begin{figure}[t]
    \centering
    \includegraphics[width=1.0\columnwidth]{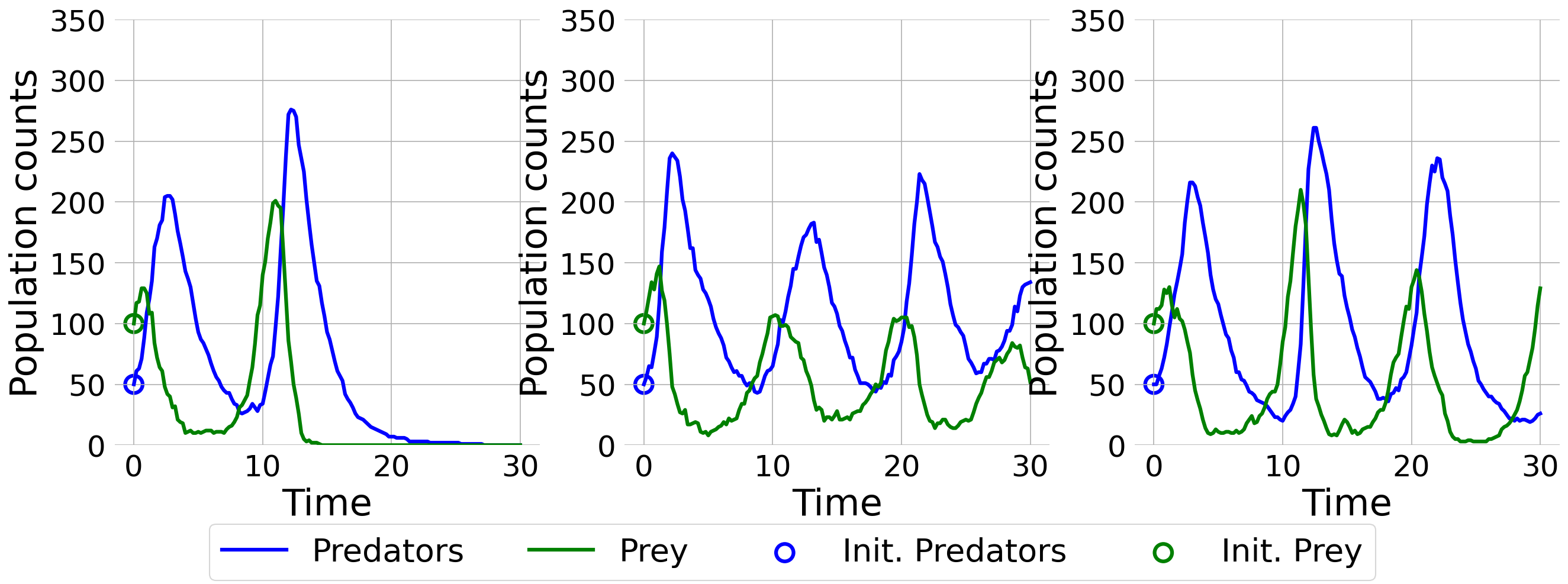}
    \caption{
    Population counts over $30$ timesteps of $3$ sample Lotka-Volterra system simulations, recording with $dt=0.2$, for $(X, Y) = (50, 100)$ and $(\theta_1, \theta_2, \theta_3, \theta_4) = (0.01, 0.5, 1.0, 0.01)$, demonstrating stochastic variability in its oscillatory behaviour due to the MJP formulation.
    }
    \label{fig:sample-lv-gt-traj}
\end{figure}

For initial populations $(X, Y) = (50, 100)$, a ground truth $\boldsymbol{\theta}_{\text{gt}} = (\theta_1, \theta_2, \theta_3, \theta_4) = (0.01, 0.5, 1.0, 0.01)$ can produce the typical oscillatory behaviour of a realistic predator-prey population system 
(\cref{fig:sample-lv-gt-traj}). 
In parameter inference for ecological modelling~\cite{wilkinson2018stochastic}, it is common practise to infer birth and death rates in the $\log$ domain using the natural logarithm (base $e$, or $\ln$)~\cite{papamakarios2016fast}. This transformation enforces positivity and improves numerical stability by smoothening the ``landscape'' that the inference algorithm has to navigate. 
Specifically, in the formulation of Lotka-Volterra as a set of nonlinear ordinary differential equations (ODEs): $\frac{dX}{dt}=\theta_1XY-\theta_2X$ and $\frac{dY}{dt}=\theta_3Y-\theta_4XY$, the dynamics are exponential, which makes $e$ the most mathematically convenient base for linking the rates to the population changes. Thus, we shall treat the Lotka-Volterra in the $\log_e$ domain, where $\ln(\theta_1, \theta_2, \theta_3, \theta_4) \approx (-4.605, -0.693, 0.0, -4.605)$. For ease of notation, we treat this $\log_e$ domain definition as the ground truth $\boldsymbol{\theta}_{\text{gt}}$. 

We further follow~\cite{papamakarios2016fast} in adopting a Markov jump process (MJP) formulation of Lotka-Volterra, in which predator and prey populations change via discrete, random events occurring in continuous time with rates derived from $\boldsymbol{\theta}$. Specifically, we draw the time to the next reaction from an exponential distribution with rate equal to the total rate $\theta_1XY + \theta_2X +\theta_3Y + \theta_4XY$. We then select a reaction $\in[1,4]$ at random, with probability proportional to its $\boldsymbol{\theta}$ rate, simulate it, and repeat.
The MJP version of the model exhibits complex and oscillatory dynamics, similar to the ODE, but with randomness replacing the dependence on initial conditions~\cite{gillespie1977exact}. 

Rather than using the raw predator-prey population timeseries as observation $\mathbf{x}$, we adopt the $9$D summary statistics proposed by~\cite{papamakarios2016fast}: 
\begin{enumerate*}[label=(\roman*)]
  \item the mean of each species' timeseries, 
  \item the log variance of each timeseries, 
  \item the autocorrelation coefficient of each timeseries at lags $1$ and $2$, and 
  \item the cross-correlation coefficient between the two timeseries. 
\end{enumerate*}

For the $\log$ domain $\boldsymbol{\theta}_{\text{gt}}$, an \plotrefblue{\emph{OK}} support is \plotrefblue{$[-5, 2] \times 4$}, while \plotrefred{$[[-3., 2.],$ $[-5., -1.5],$ $[-5., 2.],$ $[-5., 2.]]$} is a \plotrefred{misspecified} support for $\theta_1$ and $\theta_2$.
A \plotrefpurple{broad} support is \plotrefpurple{$[-6, 4] \times 4$}, a \plotrefviolet{broader} support is \plotrefviolet{$[-6, 5] \times 4$}, and the \plotrefgreen{broadest} support that gives a reasonable simulation success rate is \plotrefgreen{$[-7, 7] \times 4$}. 
Domains are underlined with reference to \Cref{fig:misspec-supp-problm-illustr}. 
Our experiments simulate $30$ steps, record $(X,Y)$ values with $dt=0.2$, and run $\max 125$ simulations (successful or failed) per iteration. 

\subsection{Inference setup}
\label{subsec:lv-lfi-setup}

We experiment with BayesSim~\cite{ramos2019bayessim}, inferring parametric MoG posteriors that tractably and interpretably group alternative hypotheses and can be directly sampled from. The number of attempted simulations per inference iteration (see \cref{alg:bsim,,alg:bsim:line:lfi-iter}) are chosen arbitrarily by the practitioner. The exact choice is usually a result of experimentation, pilot runs, and broader experience with an inference algorithm~\cite{Wilkinson2013, fearnhead2012constructing}. 

Our MDNN models $4$ Gaussians with full covariance using $3$ fully connected layers of $1024$ units each. It is trained for $1000$ epochs with the Adam optimiser, a learning rate of $1e$$-5$ and a batch size of $100$ trajectories. 
We perform $10$ inference iterations, each adding $100$ more trajectories to the training set, with parameters $\boldsymbol{\theta}$ sampled from the latest posterior.

Beyond the BayesSim experiments featured in the main sections of this paper, \Cref{appendix:snpe-misspec-support} presents a complementary study of the impact of a misspecified support on sequential neural posterior estimation, a more formally principled approach to parameter inference from the SBI literature~\cite{papamakarios2019sequential, greenberg2019icml}.

\subsection{What does the misspecified support issue look like?}
\label{subsec:problem-demo}

\begin{figure}[t]
    \centering
    \includegraphics[trim=0 90 0 0, clip, width=1.0\columnwidth]{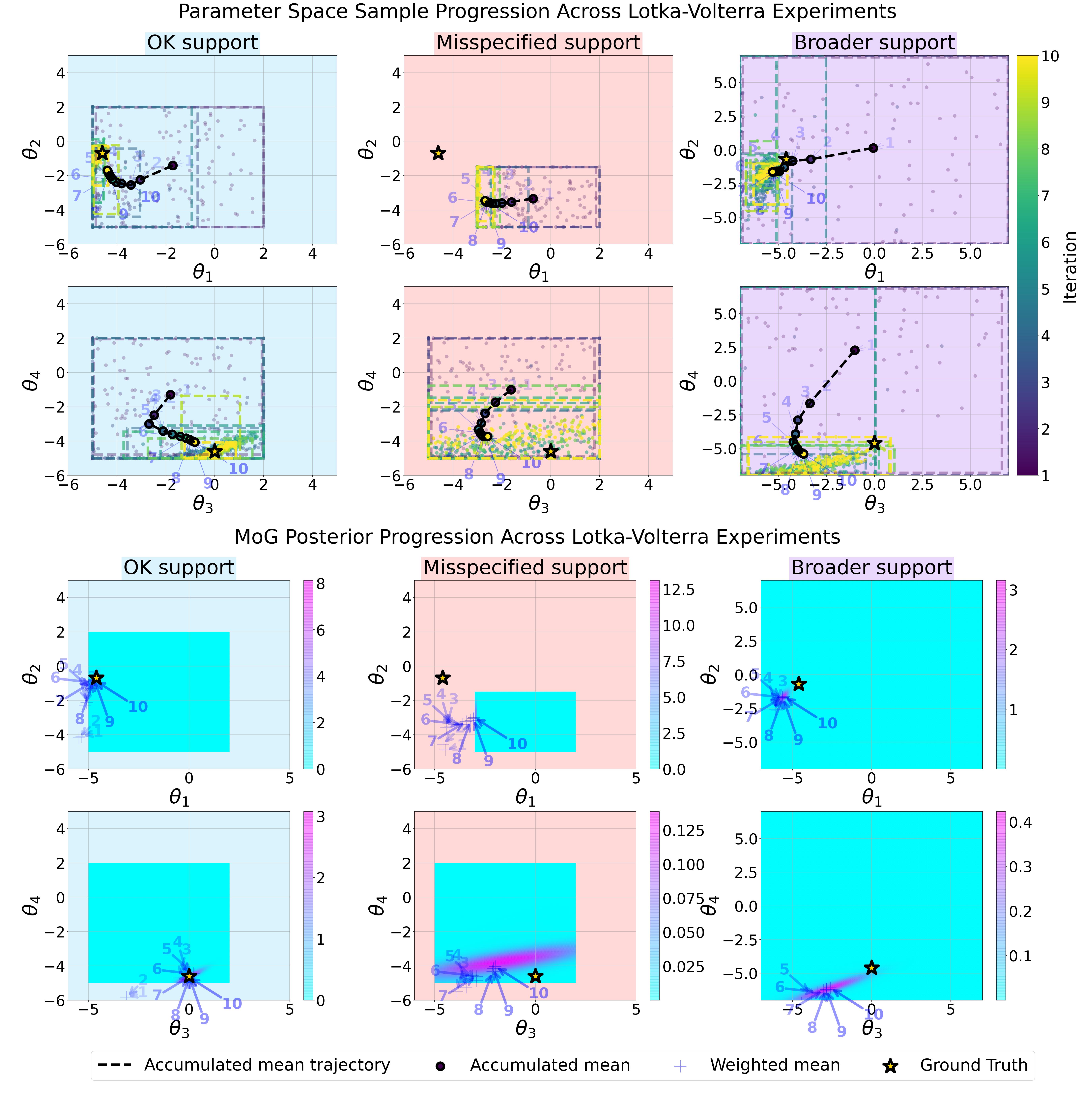}
    \caption{
    The misspecified support issue and the compensation with 
    a broader support for Lotka-Volterra. 
    We plot the prior samples of $10$ inference iterations with their bounding boxes (top) and the final MoG posterior (bottom).
    The top colorbar indicates the iteration each sample was drawn. 
    The bigger dots mark the accumulated dataset's mean in each iteration. 
    On MoG plots, the iterative MoG progression is annotated with arrows pointing to the position of the weighted mean for each iteration and colorbars quantify likelihood. 
    }
    \label{fig:misspec-supp-problm-illustr}
\end{figure}

\Cref{fig:misspec-supp-problm-illustr} shows how a misspecified support, which does not include the ground truth, leads to inaccurate and imprecise inference (\plotrefred{2nd} col.). The $\theta_1$ and $\theta_2$ support misspecification also affects the $\theta_3$ and $\theta_4$ inference, despite their subdomains being adequately specified. We also see how a wider support (\plotrefviolet{3rd} col.) not only does not guarantee stable convergence to ground truth (compared to the \plotrefblue{1st} col.), but instead hinders the learning of a reliable CDF over inference iterations.

\subsection{Why not use the widest support possible?}
\label{subsec:widest-prior}

\begin{figure}[t]
    \centering
    \includegraphics[width=1.0\columnwidth]{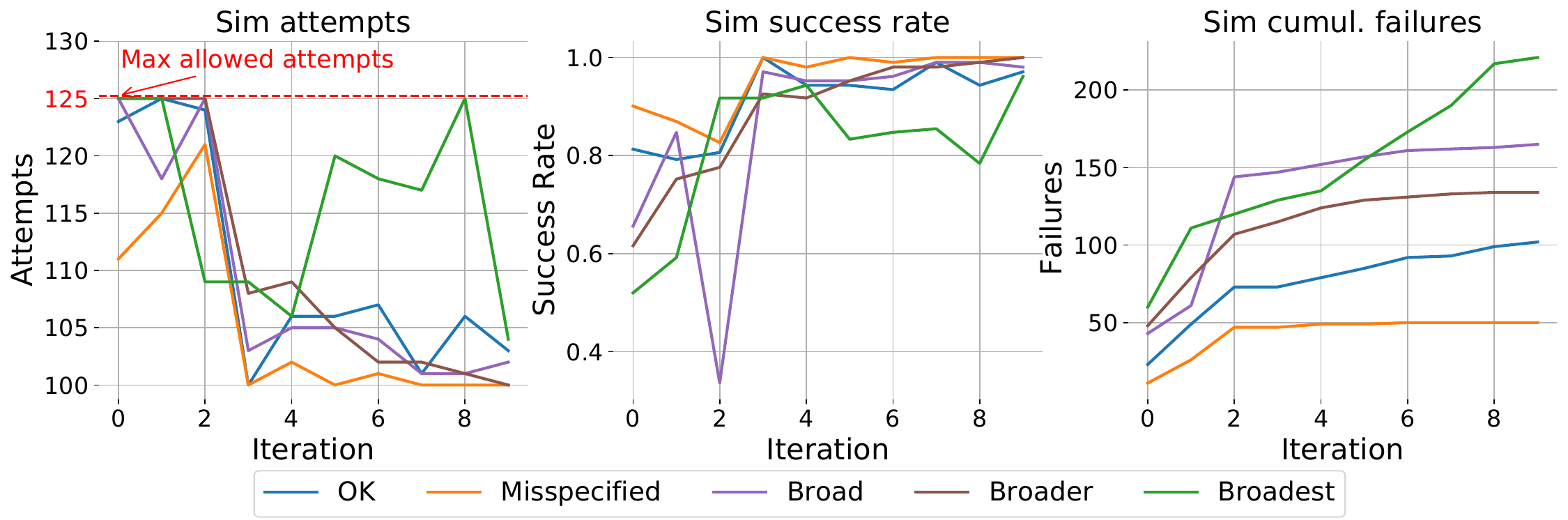}
    \caption{The reduced data efficiency of sampling over wide supports for $10$ LFI iterations for Lotka-Volterra. We plot attempted simulations (left), simulation success rate (centre) and cumulation of failed simulations (right).
    }
    \label{fig:wide-distr-problm-illustr}
\end{figure}

\Cref{fig:wide-distr-problm-illustr} shows that, when \emph{infeasible} parameterisations exist, sampling under a wider support increases simulation failure rate. 
In contrast, sampling from narrower posteriors when converging on a belief with increased certainty improves the success rate. 
Even for suboptimal \plotrefpurple{broad} and \plotrefviolet{broader} support, the success rate improves in later iterations. 
Thus, data efficiency is correlated with posterior sharpness, since both generally increase over iterations. 
Also, the narrower the support, the more likely an efficient data collection is. A wider support can reduce the success rate, requiring more simulations to converge to a posterior that may be precise but inaccurate.

\section{Posterior-driven heuristic support adaptation}
\label{sec:heur-supp-opt-methods}

Bayesian and likelihood-free methods address model misspecification by robustifying experiment selection~\cite{go2022robust} or adapting the prior~\cite{drovandi2011estimation}. However, these approaches either neglect the support or require costly redefinitions of $\Theta$ and often lack theoretical guarantees~\cite{marin2012approximate}. 
Our hypothesis is that the iterative progression of a posterior inferred under misspecified support can be treated as an information criterion that reveals both the need for adaptation and its direction~\cite{di2023active}. 
We therefore introduce three heuristic extensions to BayesSim (\cref{alg:bsim}, \cref{alg:bsim:hda-extension}). 
Each heuristic deterministically answers: 
\emph{``Given $\hat{p}(\boldsymbol{\theta} \mid \mathbf{x} = \mathbf{x}^r)$, how should the support be adapted to improve subsequent inference?''}. 

We treat support adaptation as expansion; specifically, as \emph{window stretching}~\cite{akkaya2019solving}, not \emph{window sliding}~\cite{mehta2021user}: instead of shifting the sampling window, we iteratively accumulate new samples using the latest posterior (\cref{alg:bsim}, \cref{alg:bsim:line:dataset-update}). 

\subsection{BayesSim-EDGE: expanding bounds via edge mass}
\label{subsec:bsim-edge}

\begin{algorithm}[!t]
\caption{The \textbf{EDGE} heuristic}
\label{alg:bsim-edge}
\begin{algorithmic}[1]
    \Function{\textcolor{cyan!60!black}{EDGE}}{$\Theta_0, \Phi, \delta, \tau, \eta$}
    \For{$d = 1$ to $D$}
        \Comment{Iterate param. dims}
        \State $r^{(d)} = \lvert \theta_{\max}^{(d)} - \theta_{\min}^{(d)} \rvert$
        \Comment{Range size}
        \State $\Delta_L^{(d)} \gets [\theta_{\min}^{(d)}, \theta_{\min}^{(d)} + \delta ]$
        \Comment{Left edge zone}
        \State $\Delta_R^{(d)} \gets [\theta_{\max}^{(d)} - \delta, \theta_{\max}^{(d)}]$
        \Comment{Right edge zone}
        \State Accum. masses $M_L^{(d)}$ in $\Delta_L^{(d)}$ \& $M_R^{(d)}$ in $\Delta_R^{(d)}$
        \If{$M_L^{(d)} > \tau \And (\theta_{\min}^{(d)} - \eta r^{(d)}) \in \Phi$} \label{alg:bsim-edge:left-edge-criterion}
            \State $\theta_{\min}^{(d)} \gets \theta_{\min}^{(d)} - \eta r^{(d)}$ \label{alg:bsim-edge:left-edge-expansion}
            \Comment{Expand left}
        \EndIf
        \Comment{\emph{if} $\notin \Phi$ \emph{then} $\theta_{\min}^{(d)} \gets \Phi_{\min}^{(d)}$}
        \If{$M_R^{(d)} > \tau \And (\theta_{\max}^{(d)} + \eta r^{(d)}) \in \Phi$} \label{alg:bsim-edge:right-edge-criterion}
            \State $\theta_{\max}^{(d)} \gets \theta_{\max}^{(d)} + \eta r^{(d)}$ \label{alg:bsim-edge:right-edge-expansion}
            \Comment{Expand right}
        \EndIf
        \Comment{\emph{if} $\notin \Phi$ \emph{then} $\theta_{\max}^{(d)} \gets \Phi_{\max}^{(d)}$}
    \EndFor
    \State \Return $\boldsymbol{\theta}_{\min}, \boldsymbol{\theta}_{\max}$
    \EndFunction
\end{algorithmic}
\end{algorithm}

BayesSim-EDGE (\textbf{E}dge-\textbf{D}riven \textbf{G}aussian \textbf{E}xpansion) expands the support bounds (\emph{`edges'}) based on the accumulation of posterior mass near them (\cref{alg:bsim-edge}). The motivation is to expand the parameter search space when the learnt posterior exhibits a significant mass close to existing bounds, suggesting that the true parameters may lie beyond them.

We begin with \emph{standard BayesSim assumptions}, meaning a proposal prior $\Tilde{p}$ from which we sample the parameters $\boldsymbol{\theta}$ and a policy $\pi_{\boldsymbol{\beta}_0}$ to be used for data collection both in sim and in real (for $\mathbf{x}^r$). We assume the proposal prior to be uniform and we expand our formulation to concretely define the initial support as $\Theta_0 = [\boldsymbol{\theta}_{\min}, \boldsymbol{\theta}_{\max}]$, or, compactly, $\Tilde{p}_{(\Theta_0)}$.

EDGE is configurable through its hyperparameters: 
the edge mass threshold $\tau$, the edge zone fraction $\delta$, and the edge expansion factor $\eta$. 
Specifically, $\delta$ defines the size of the area within which accumulated probability mass is considered \emph{near} the edge, and thus the sensitivity of the expansion criterion, which compares the edge mass with a threshold $\tau$. We denote the left and right edge zones ($\min$ and $\max$ side) as $\Delta_L^{(d)}$ and $\Delta_R^{(d)}$ for a dimension $d$ of $\boldsymbol{\theta}$, and their accumulated masses as $M_L^{(d)}$ and $M_R^{(d)}$. 
If an edge's mass exceeds $\tau$ and the expansion remains within $\Phi$ (\cref{alg:bsim-edge:left-edge-criterion,,alg:bsim-edge:right-edge-criterion}), we expand the respective domain boundary by a factor of $\eta$ (\cref{alg:bsim-edge:left-edge-expansion,,alg:bsim-edge:right-edge-expansion}). If the expansion would extend beyond the extremum of the respective $\Phi$ limit, then this becomes the new boundary.

EDGE explores beyond the initial parameter ranges if the inferred posterior (\cref{alg:bsim:line:cdf-update}) confidently points towards the edges of the domain. 
This \emph{confidence} is tuned through $\tau$ and $\delta$. Thus, within $N_{\text{LFI}}$ iterations, we can converge to an adapted support $\Theta_{N_{\text{LFI}}}$ in conjunction with a posterior $\hat{p}$. 
Conceptually, EDGE assumes that if there is an accumulation of $\hat{p}$ mass near an edge, then we should extend the support in this direction. This rule treats posterior mass accumulation as a geometric indication that there is more value in exploring beyond the corresponding $\Theta$ boundary.

In this paper, all heuristic hyperparameters are denoted as scalars, 
but can also be vectors, e.g. $\tau \rightarrow \boldsymbol{\tau}$. This enables custom rules for each $\boldsymbol{\theta}$ dimension, e.g. having different edge mass thresholds for the DLO Young's modulus and length. 

\subsection{BayesSim-MODE: mode-shift-based expansion}
\label{subsec:bsim-mode}

\begin{algorithm}[!t]
\caption{The \textbf{MODE} heuristic}
\label{alg:bsim-mode}
\begin{algorithmic}[1]
    \Function{\textcolor{cyan!60!black}{MODE}}{$\Theta_0, \Phi, \nu_{\text{TH}}, \rho, \tau, \eta$}
    \While{$i > 0 \And d < D$}
        \Comment{Iterate param. dims}
        \State $r^{(d)} = \lvert \theta_{\max}^{(d)} - \theta_{\min}^{(d)} \rvert$
        \Comment{Range size}
        \State $W_L^{(d)} \gets 0$; $W_R^{(d)} \gets 0$
        \Comment{MoG weights accum.}
        \ForEach{Gaussian $k \in$ MoG}
            \State $\nu_{i,k}^{(d)} \gets \mu_{i,k}^{(d)} - \mu_{i-1,k}^{(d)}$ \label{alg:bsim-mode:mode-shift}
            \Comment{Mode shift}
            \State $z_{k}^{(d)} \gets \frac{(\mu_{i,k}^{(d)} - \theta_{\min}^{(d)})}{r^{(d)}}$ \label{alg:bsim-mode:mode-bounds-proximity}
            \Comment{Std. bound proximity}
            \If{$\nu_{i,k}^{(d)} < -\nu_{\text{TH}}$ $\And$ $|z_{k}^{(d)}| < \rho$}
                \State $W_L^{(d)} \gets W_L^{(d)} + w_{i,k}$ \label{alg:bsim-mode:left-weight-accum}
                \Comment{Accum. left}
            \ElsIf{$\nu_{i,k}^{(d)} > \nu_{\text{TH}}$ $\And$ $|1 - z_{k}^{(d)}| < \rho$}
                \State $W_R^{(d)} \gets W_R^{(d)} + w_{i,k}$ \label{alg:bsim-mode:right-weight-accum}
                \Comment{Accum. right}
            \EndIf
        \EndFor
        \If{$W_L^{(d)} > \tau \And (\theta_{\min}^{(d)} - \eta r^{(d)}) \in \Phi$} \label{alg:bsim-mode:left-edge-criterion}
            \State $\theta_{\min}^{(d)} \gets \theta_{\min}^{(d)} - \eta r^{(d)}$ \label{alg:bsim-mode:left-edge-expansion}
            \Comment{Expand left}
        \EndIf
        \Comment{\emph{if} $\notin \Phi$ \emph{then} $\theta_{\min}^{(d)} \gets \Phi_{\min}^{(d)}$}
        \If{$W_R^{(d)} > \tau \And (\theta_{\max}^{(d)} + \eta r^{(d)}) \in \Phi$} \label{alg:bsim-mode:right-edge-criterion}
            \State $\theta_{\max}^{(d)} \gets \theta_{\max}^{(d)} + \eta r^{(d)}$ \label{alg:bsim-mode:right-edge-expansion}
            \Comment{Expand right}
        \EndIf
        \Comment{\emph{if} $\notin \Phi$ \emph{then} $\theta_{\max}^{(d)} \gets \Phi_{\max}^{(d)}$}
        \State $d \gets d + 1$
    \EndWhile
    \State \Return $\boldsymbol{\theta}_{\min}, \boldsymbol{\theta}_{\max}$
    \EndFunction
\end{algorithmic}
\end{algorithm}

BayesSim-MODE (\textbf{M}ode-\textbf{O}riented \textbf{D}omain \textbf{E}xpansion) expands the support based on the position shift in the posterior modes between consecutive inference iterations in conjunction with their proximity to the boundaries (\cref{alg:bsim-mode}). The intuition is that high-confidence mode shift toward the boundaries of the current support may indicate that the true parameters lie beyond them. In contrast to EDGE, which checks a static edge mass, MODE uses temporal information. 

Similarly to EDGE, the standard BayesSim assumptions apply. We also have MODE's hyperparameters: 
the mode shift threshold $\nu_{\text{TH}}$, the bounds proximity threshold $\rho$, the MoG weights summary threshold $\tau$, the bounds expansion factor $\eta$, and the physically feasible domain limits $\Phi$. 
In contrast to EDGE, MODE computes the shift of the MoG posterior modes 
$\boldsymbol{\mu}_{i}$ compared to the previous iteration $\boldsymbol{\mu}_{i-1}$ (\cref{alg:bsim-mode:mode-shift}) along with their normalised proximity to the bounds (\cref{alg:bsim-mode:mode-bounds-proximity}).
Thus, we do not check for $\Theta$ adaptation at the end of the first inference iteration. In all subsequent iterations, the MoG details of the previous iteration are accessible through the reference prior.

We accumulate the mixture coefficients (weights) of each mode that satisfies the shift and proximity thresholds for the left or right bound (\cref{alg:bsim-mode:left-weight-accum,,alg:bsim-mode:right-weight-accum}) of a parameter $d$. The accumulated weights $W_L^{(d)}$, $W_R^{(d)}$ are used for the expansion criteria (\cref{alg:bsim-mode:left-edge-criterion,,alg:bsim-mode:right-edge-criterion}). 
Similar to EDGE, if $\tau$ has been exceeded, we expand it by a factor of $\eta$ (\cref{alg:bsim-mode:left-edge-expansion,,alg:bsim-mode:right-edge-expansion}), or as far as $\Phi$ allows. 

MODE, therefore, trades the EDGE's requirement of configuring the probability mass thresholds with configuring the mode-shift thresholds.
Conceptually, MODE assumes that if there is a shift ($\nu_{\text{TH}}$, $\rho$) of high-certainty MoG modes (mixture weights, $\tau$) towards an edge, then we should extend the support in this direction. This rule treats the posterior mode shift as a temporal sign that there is more value in exploring beyond the corresponding $\Theta$ boundary.

\subsection{BayesSim-CENTRE: centring-based adaptive bounds}
\label{subsec:bsim-centre}

\begin{algorithm}[!t]
\caption{The \textbf{CENTRE} heuristic}
\label{alg:bsim-centre}
\begin{algorithmic}[1]
    \Function{\textcolor{cyan!60!black}{CENTRE}}{$\Theta_0, \Phi$}
    \For{$d = 1$ to $D$}
        \Comment{Iterate param. dims}
        \State $r^{(d)} = \lvert \theta_{\max}^{(d)} - \theta_{\min}^{(d)} \rvert$
        \Comment{Range size}
        \State $\mu^{(d)} \gets$ $\hat{p}$ modes weighted mean \label{alg:bsim-centre:line:weighted-mean}
        \Comment{New \emph{centre}}
        \State ${\theta_{\min}^{(d)}}^{\prime} \gets \mu^{(d)} - \frac{r^{(d)}}{2}$ \label{alg:bsim-centre:left-edge-candidate}
        \Comment{$\min$ candidate}
        \State ${\theta_{\max}^{(d)}}^{\prime} \gets \mu^{(d)} + \frac{r^{(d)}}{2}$ \label{alg:bsim-centre:right-edge-candidate}
        \Comment{$\max$ candidate}
        \If{${\theta_{\min}^{(d)}}^{\prime} \notin \Phi$} \label{alg:bsim-centre:left-edge-criterion}
            \Comment{Check if $\min$ feasible}
            \State ${\theta_{\min}^{(d)}}^{\prime} \gets \Phi_{\min}^{(d)};\;{\theta_{\max}^{(d)}}^{\prime} \gets {\theta_{\min}^{(d)}}^{\prime} + r^{(d)}$
        \EndIf
        \If{${\theta_{\max}^{(d)}}^{\prime} \notin \Phi$} \label{alg:bsim-centre:right-edge-criterion}
            \Comment{Check if $\max$ feasible}
            \State ${\theta_{\max}^{(d)}}^{\prime} \gets \Phi_{\max}^{(d)};\;{\theta_{\min}^{(d)}}^{\prime} \gets {\theta_{\max}^{(d)}}^{\prime} - r^{(d)}$
        \EndIf
        \State $\theta_{\min}^{(d)} \gets {\theta_{\min}^{(d)}}^{\prime};\;\theta_{\max}^{(d)} \gets {\theta_{\max}^{(d)}}^{\prime}$
        \Comment{Shift $\Theta$ bounds}
    \EndFor
    \State \Return $\boldsymbol{\theta}_{\min}, \boldsymbol{\theta}_{\max}$
    \EndFunction
\end{algorithmic}
\end{algorithm}

BayesSim-CENTRE moves the support boundaries such that in each inference iteration the weighted mean of the posterior modes is in the centre of the adapted support (\cref{alg:bsim-centre}). 
The intuition is that this zero-assumption approach will enable adapting the support to better match the current estimate of $\boldsymbol{\theta}$.

Thus, beyond the standard BayesSim assumptions, we now only have physically feasible domain limits $\Phi$. 
In each inference step, following the posterior $\hat{p}$ update, we compute the weighed mean $\mu^{(d)}$ of the $\hat{p}$ modes for each dimension $d$ of the support (\cref{alg:bsim-centre:line:weighted-mean}). We use $\mu^{(d)}$ to compute the candidate bounds of the adapted support 
(\cref{alg:bsim-centre:left-edge-candidate,,alg:bsim-centre:right-edge-candidate}). We then evaluate these candidate bounds for their physical feasibility (\cref{alg:bsim-centre:left-edge-criterion,,alg:bsim-centre:right-edge-criterion}), and adjust them accordingly. 
The result is the adapted support. 

CENTRE requires less knowledge of LFI performance in a given task, since it does not have any task-specific hyperparameters. 
Also, it does a \emph{window sliding} support adaptation, compared to EDGE and MODE \emph{window stretching}. 
Conceptually, CENTRE assumes that the true parameters lie approximately at the mean of the peaks of the MoG components. This treats the positions and weights of the MoG modes as geometric indications of the range of the optimal search space.

\section{Proof-of-concept with Lotka-Volterra}
\label{sec:lv-poc-experiment}

We compare the standard BayesSim~\cite{ramos2019bayessim} and our heuristic variants.
EDGE is parameterised with $\delta=0.1$, $\tau=0.005$ and $\eta=0.2$, and MODE with $\nu_{\text{TH}}=0.01$, $\rho=0.4$, $\tau=0.05$ and $\eta=0.2$. 
For a sensitivity analysis of different EDGE and MODE hyperparameter choices, see \cref{appendix:heur-sensitivity}.
We use an MDNN for the CDF approximation, configured as in \Cref{subsec:lv-lfi-setup}.

\begin{figure}[t]
    \centering
    \includegraphics[trim=0 45 0 0, clip, width=1.0\columnwidth]{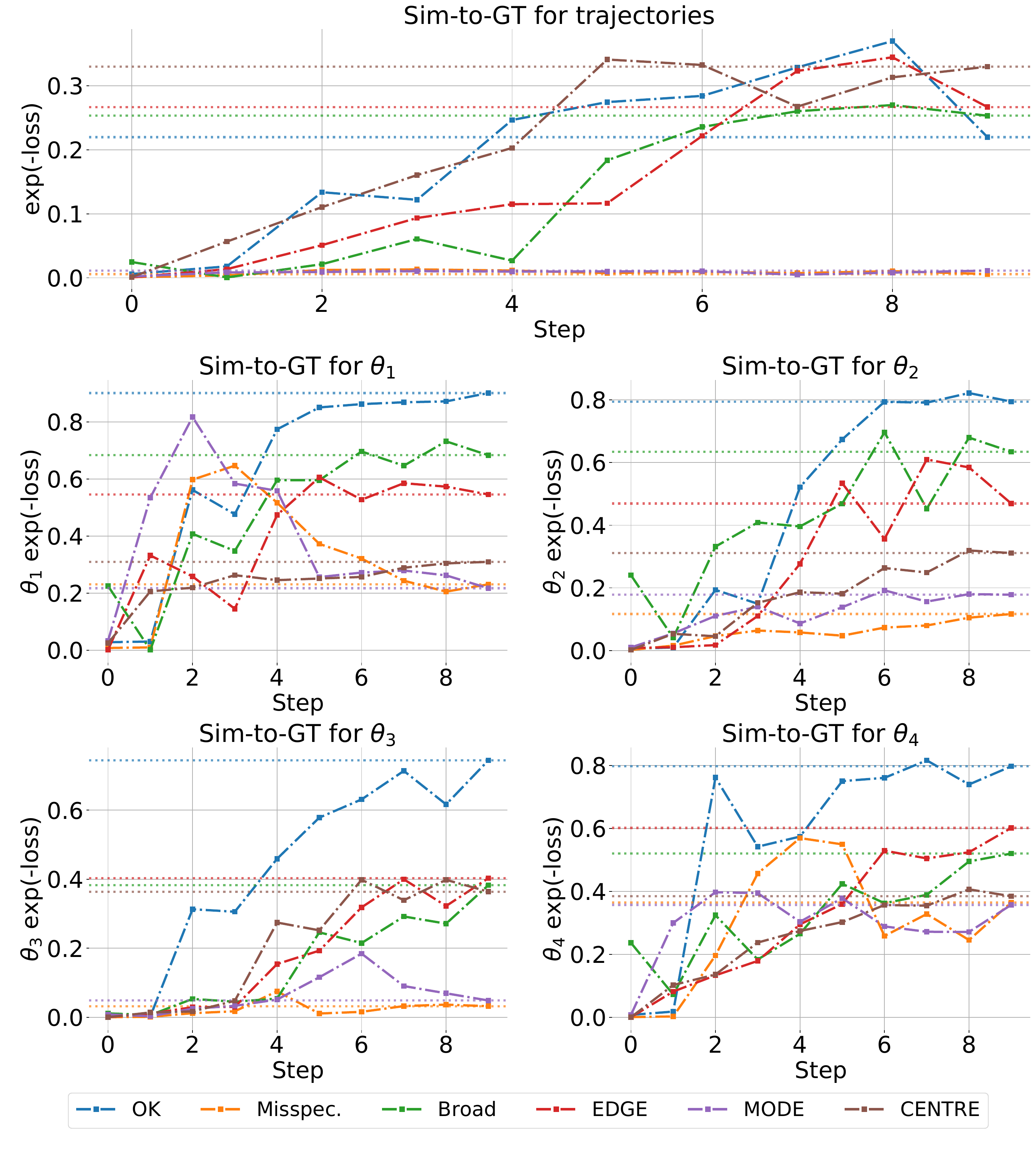}
    \caption{Lotka-Volterra Sim2Sim inference exponential of negative loss for trajectories (top) and parameters $\boldsymbol{\theta}$ (bottom grid) over $10$ iterations.
    }
    \label{fig:lv-res-curves}
\end{figure}

\begin{figure}[t]
    \centering
    \includegraphics[width=1.0\columnwidth]{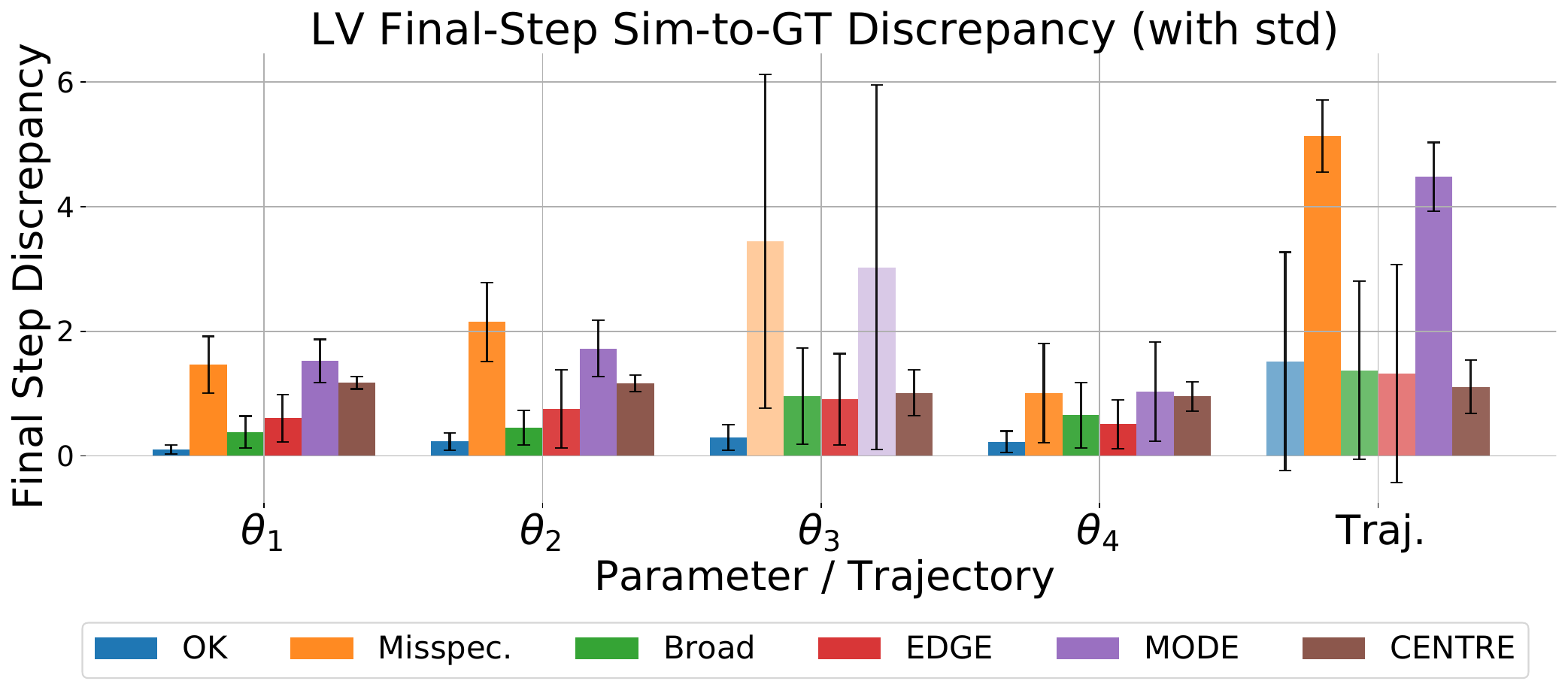}
    \caption{
    Final inference step discrepancy (i.e., at \nth{10} \emph{iteration}) from ground-truth parameters and trajectory for the Lotka-Volterra Sim2Sim experiments.
    }
    \label{fig:lv-discrepancy-bars}
\end{figure}

\Cref{fig:lv-res-curves} shows the LFI results in Lotka-Volterra for different configurations of the inference problem. 
We consider a physically feasible domain $\Phi = [-6, 4] \times 4$, which is the broad support from \cref{subsec:lv-exper}.
We plot the exponential of the negative of all resulting loss ($\mathcal{L}$) values ($\exp(-\alpha \mathcal{L})$, with $\alpha=1.0$), compressing higher values and extending lower ones 
(as $\mathcal{L} \rightarrow 0$). This gives a fine-grained insight into performance differences among our heuristic support adaptation methods. 
The top plot shows the transformed 
loss 
of the simulated $X,Y$ population trajectory to the ground truth, averaged for $20$ $\boldsymbol{\theta}$ samples from each iteration's posterior and simulating for each. 
Original loss is measured as the Euclidean distance.
The bottom grid shows the transformed 
losses 
of the inferred $\boldsymbol{\theta}$ to $\boldsymbol{\theta}_{\text{gt}}$, averaged for $20$ samples from each iteration's posterior. 
Original loss is measured as the absolute difference. 

With a good support (\emph{OK}), the Sim2Sim inference consistently approximates the ground-truth $X,Y$ population trajectories. 
CENTRE and EDGE tend to perform better than using broad support. 
Although CENTRE is better in terms of trajectory proximity to ground truth, the per-parameter plots show that EDGE is more robust in per-parameter prediction accuracy. 
MODE underperforms, with its temporal shift heuristic being difficult to tune, especially given the task's stochastic dynamics. 
Thus, it is almost as inaccurate as the misspecified support inference. These results also motivate a more thorough exploration of how each physical parameter's value influences the resulting population trajectories. 
Qualitatively, our results suggest that the inference of 
$\theta_2$ (a predator dies) and $\theta_4$ (a prey dies) is better correlated with ground-truth trajectory approximation, followed by $\theta_3$ (a prey is born). 

These insights are supported by \Cref{fig:lv-discrepancy-bars}, which illustrates the discrepancy of the inferred $\boldsymbol{\theta}$ after $10$ inference iterations, averaged from $20$ posterior samples with the error-bars being the standard deviations. 
EDGE leads to the smallest overall discrepancy in predicted parameters and resulting trajectories. A broad support may occasionally surpass EDGE 
in per-parameter prediction accuracy, but practitioners need to manage data efficiency and simulation feasibility issues (\cref{subsec:widest-prior}).

Overall, EDGE's success suggests that a probability mass heuristic is a more robust approach to support adaptation in stochastic dynamical systems, corroborated by the sensitivity analysis of \cref{appendix:heur-sensitivity}.
Still, it requires careful hyperparameter tuning, which can draw on prior task experience obtained through manual experimentation and pilot runs~\cite{Wilkinson2013, fearnhead2012constructing}; 
further investigation is beyond the scope of this work. 
Among the other heuristic variants, CENTRE seems more useful than MODE, likely due to the difficulty of tuning the mode-shift tracking hyperparameters for a stochastic iterative inference method. 
However, despite making zero assumptions, CENTRE can also struggle for certain tasks, as Bayesian LFI inherently tends to place probability mass on the most likely candidates within the given parameter space, which can negatively impact a centring-based sliding window approach.

\section{Real2Sim2Real for visuomotor DLO whipping}
\label{sec:r2s2r-dlo-whip-exper}

\begin{figure}[t]
    \centering
    \includegraphics[width=1.0\columnwidth]{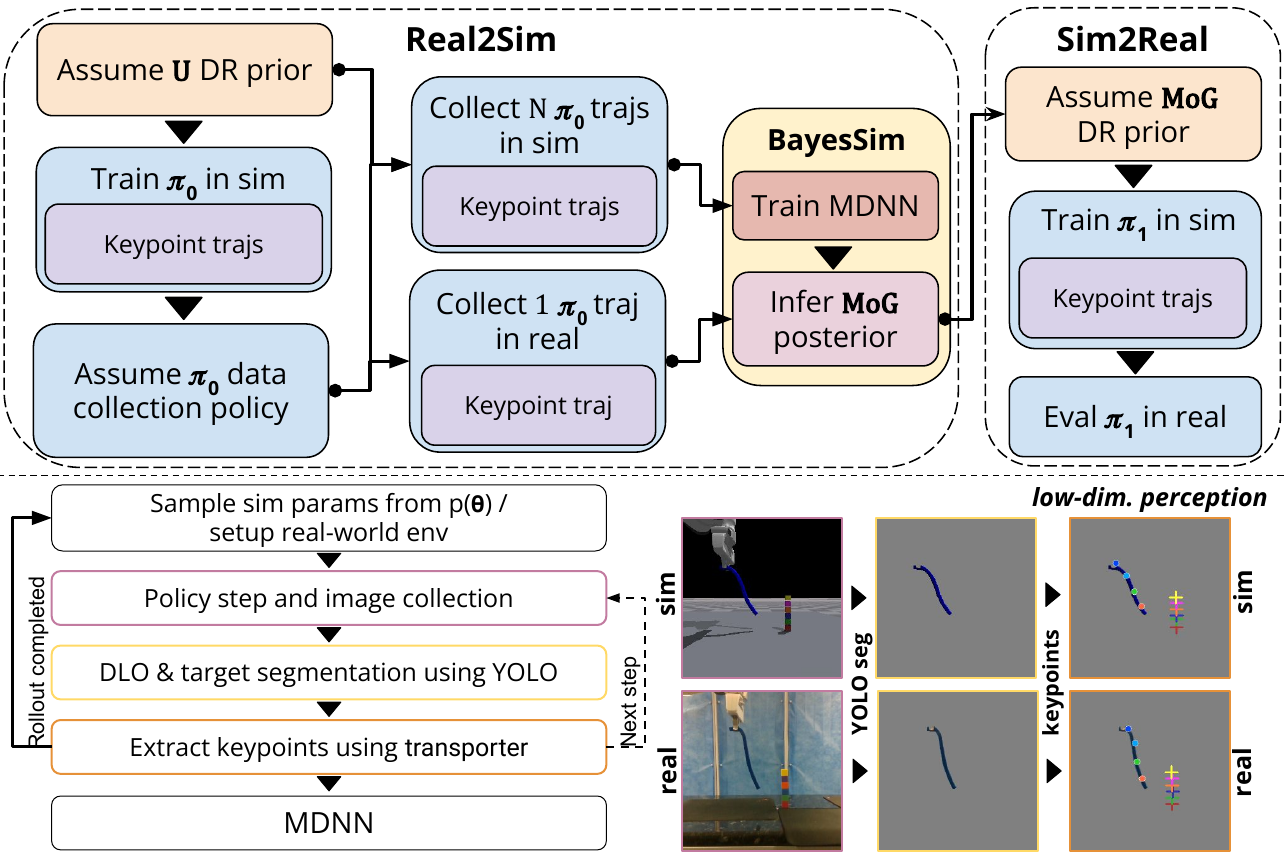}
    \caption{We use a Real2Sim2Real approach to visuomotor DLO whipping (top). We infer the posterior $\hat{p}(\boldsymbol{\theta})$ (Real2Sim) and use it for DR to train a PPO agent. We deploy and evaluate the sim-trained policy in the real world (Sim2Real). 
    In each policy step we infer the DLO and stacked cubes keypoints from segmentation images and use them as state input for the MDNN (bottom). 
    }
    \label{fig:r2s2r-n-perc-workflow}
\end{figure}

Following our Lotka-Volterra results, we use BayesSim-EDGE to study a visuomotor DLO whipping task with a sparse outcome-dependent reward that guides the entire DLO body~\cite{yu2025generalizable} toward a stack of cubes.
We use the Real2Sim2Real framework of~\cite{kamaras2025distributional} for vision-based DLO manipulation (\cref{fig:r2s2r-n-perc-workflow}).

\subsection{Task overview}
\label{subsec:task-setup}

\subsubsection{Initialisation}
\label{subsubsec:task-init}
A robot arm picks up the DLO from a designated position $\mathbf{x}_0$ on the table by grasping it near one of its tips and raising it to a designated height $h_0$. Both $\mathbf{x}_0$ and $h_0$, as well as the initial DLO pose, remain fixed for all the simulation and real experiments. 
The execution of the whipping policy begins once the DLO is raised to $h_0$. 

\subsubsection{Visuomotor elements}
\label{subsubsec:task-visuomotor}
We use two dedicated keypoint clusters to track the DLO and the stack of cubes in the 2D pixel space of an RGB image.
In both parameter inference and policy learning and deployment, we position-control the end effector (EEF) by commanding its Cartesian pose. Our simulator uses the IsaacGym attractors implementation, whereas in the real world we use a Cartesian impedance controller~\cite{luo2024serl}.

\subsection{Simulation setup}
\label{subsec:sim-setup}

We use IsaacGym with FleX physics~\cite{makoviychuk2021isaac} to simulate 
a Franka Emika Panda 7-DoF robot arm with a parallel gripper on a tabletop. 
We also have a blue DLO, simulated as a tetrahedral grid with the corotational finite element method. 
The whipping targets are a stack of $6$ distinctly coloured cubes: $[$\setulcolor{red}\ul{red}, \setulcolor{green}\ul{green}, \setulcolor{blue}\ul{blue}, \setulcolor{orange}\ul{orange}, \setulcolor{purple}\ul{purple}, \setulcolor{yellow}\ul{\emph{yellow}}$]$, in bottom-up order, the \setulcolor{yellow}\ul{\emph{yellow}} \emph{at the top} being the \emph{ideal target} for \emph{max reward}.

In all inference experiments, we have an initial support $\Theta_0$ and a physically feasible domain $\Phi$, with $\Theta_0 \subseteq \Phi$ and similarly each inference iteration's (index $i$) support being $\Theta_i \subseteq \Phi$. When training a policy with uniform DR ($\text{PPO-}\mathit{U}$ in~\cref{subsec:policy-learn-sim2real-res}), we sample over $\Theta_0$. When training a policy using an inferred MoG for DR, we implicitly sample the support in which we converged by the last LFI iteration (\cref{alg:bsim}, \cref{alg:bsim:line:posterior-policy-train}).

In $\Theta_0$, the DLO is sampled in $[1e3, 5e4]\,\si{\pascal}$ for Young's modulus ($E$) and in $[195, 305]\,\si{\milli\meter}$ for length ($l$).  
In $\Phi$, Young's is sampled in $[0.5e3, 5e5]\,\si{\pascal}$ and length in $[50, 350]\,\si{\milli\meter}$. During policy training, we also uniformly randomise the DLO density ($\rho_{dlo}$) in $[50, 100]\,\si{\kilogram\per\cubic\meter}$, its Poisson's ratio ($\nu$) in $[0.3, 0.5]$, the cube side lengths ($s$) in $[20, 30]\,\si{\milli\meter}$ and their density ($\rho_{cube}$) in $[80, 120]\,\si{\kilogram\per\cubic\meter}$. 
Given the above, \emph{median} of $\Theta_0$ for $(E, l, \rho_{dlo}, \nu, s, \rho_{cube})$ is 
\scalebox{0.95}{$\mu = (2.5e4\,\si{\pascal}, \, 250\,\si{\milli\meter}, \, 75\,\si{\kilogram\per\cubic\meter}, \, 0.35, \, 25\,\si{\milli\meter}, \, 100\,\si{\kilogram\per\cubic\meter})$}.  

\subsection{Real-world setup}
\label{subsec:real-setup}

Our real-world setup closely replicates the simulation.

\subsubsection{Camera details}
\label{subsubsec:camera-details}
We collect visual observations at $60$ fps with a RealSense D435i camera that views the workspace from the right side. To avoid a more elaborate sim and real camera calibration, we position the real camera so that the captured images approximate the respective sim images~\cite{matl2020inferring, antonova2022bayesian}.

\subsubsection{DLO details}
\label{subsubsec:dlo-details}
We manufacture $4$ blue DLOs of varied length using silicone polymers of various Shore hardness~\cite{liao2020ecoflex}. 
As in simulation, the real DLOs are grid-shaped, with negligible height and width at $15\,\si{\milli\meter}$ each. 
\Cref{tab:dlos} lists the DLO indexes, parameterisations, relative hardness, and mass, sorted primarily by increasing length and then by increasing softness (as \emph{medium soft $\rightarrow$ soft $\rightarrow$ extra soft}).
We will be referencing DLOs using their index, or as ``$\langle \text{Length} \rangle ; \langle \text{Shore Hardness} \rangle$''.

\subsubsection{Cube details}
\label{subsubsec:cube-details}
The real cubes are made of lightweight foam, weigh $1.54\,\si{\gram}$ and have side length $\approx 25\,\si{\milli\meter}$. Their colours and stacking order are as in simulation. 

\begin{table}[t]
\caption{Real DLO indexes and parameterisations}
\label{tab:dlos}
\centering
\begin{tabular}{c c c c c} 
    \textbf{DLO idx.} & 0 & 1 & 2 & 3 \\
    \midrule
    \textbf{Length} (\si{\milli\meter}) & $200$ & $200$ & $270$ & $290$  \\
    \midrule
    \textbf{Shore} & A-40 & 00-20 & 00-50 & 00-20 \\
    \textbf{Hardness} & \emph{(medium soft)} & \emph{(extra soft)} & \emph{(soft)} & \emph{(extra soft)} \\ 
    \midrule
    \textbf{Mass} (\si{\gram}) & $47.47$ & $42.94$ & $57.8$ & $63.94$ 
\end{tabular}
\end{table}

\subsection{Visual perception setup}
\label{subsec:vis-perc-setup}

\subsubsection{Segmentation images}
\label{subsubsec:segm-details}
For efficiency, we focus our observations on the DLO and the cubes using real-time segmentation images (\cref{fig:r2s2r-n-perc-workflow}). For this, we fine-tune the segmentation version of YOLOv11~\cite{redmon2016you} for $64$ epochs. 
We use a dataset of $311$ manually labelled images featuring real and simulated DLO physical interactions, $128$ of which are dedicated to DLO whipping. We pre-process with auto-orientation and resizing to $256\times256$, and augment with random Gaussian blur and salt-and-pepper noise, for a final dataset of $745$ images. 

\subsubsection{Keypoints}
\label{subsubsec:kps-details}
We further reduce the dimensions of our observations by using keypoints to track the DLO and the cubes. 
For the DLO, we infer $4$ keypoints in real time over its segmentation image 
with a \emph{transporter} model~\cite{kulkarni2019unsupervised, li2020causal}, trained for $50$ epochs over $1500$ random policy rollouts of simulated visuomotor DLO reaching~\cite{kamaras2025distributional} with $\boldsymbol{\theta}$ sampled from a uniform prior.
This dataset also includes a small number of real-world reaching rollouts. 
For the $6$ cubes, we treat the centroids of their individual segmentation masks as their keypoints.
These $10 \times 2$D keypoints are our visual observation vectors.

\subsection{Proprioception \& control setup}
\label{subsec:ctrl-setup}

\subsubsection{Observation \& action vectors}
\label{subsubsec:obs-n-act-vecs}
We control the EEF through Cartesian pose commands ($7$D vectors). 
The EEF moves in $2$D by controlling the $x$ and $z$ axes' deltas.
Thus, the policy actions are $2$D $\langle dx, dz \rangle$ vectors that we sample in the $[-0.06, 0.06]\,\si{\meter}$ range for smooth EEF transitions. Our $28$D observations are constructed by concatenating the EEF's $\langle x, z \rangle$ $2$D position (proprioception) and the $26$D visual observation, which the flattened $10 \times 2$D vector of the $4$ DLO and $6$ cube keypoints 
concatenated with the $6$D boolean vector that tracks whether a cube has been hit in the current episode.

\subsubsection{Safety limits}
\label{subsubsec:eef-workspace}
We restrict the EEF motion within the $\langle x, y, z \rangle \in \langle [0.275, 0.6], [-0.1, 0.1], [0.1, 0.5] \rangle\,\si{\meter}$ world frame coordinates, with the robot arm based on $(0, 0, 0)$. Whenever the EEF leaves this designated workspace, the episode ends as a failure with a reward of $-1$. 

\subsection{Real2Sim parameter inference setup}
\label{subsec:real2sim-bayessim}

We define a physical parameter vector $\boldsymbol{\theta} = \langle l, E \rangle$ and infer a joint posterior $\hat{p}(\boldsymbol{\theta})$, following \Cref{alg:bsim}. 
We use the standard BayesSim algorithm~\cite{ramos2019bayessim} and our EDGE variant (\cref{alg:bsim-edge}). 
EDGE is parameterised with $\delta=0.1$, $\boldsymbol{\tau}=\langle 0.003, 0.005 \rangle$ for DLOs-$\{0,1,3\}$ or $\tau=0.005$ for DLO-$2$, and $\eta=0.2$.
We use an MDNN for the CDF approximation, configured as in \Cref{subsec:lv-lfi-setup}, but with a batch size of $10$ trajectories.
We approximate the posterior $\hat{p}(\boldsymbol{\theta})$ through $15$ iterations, each augmenting the training set with $100$ more trajectories, whose parameters $\boldsymbol{\theta}$ have been sampled using the latest posterior.

For all simulated and real trajectories $\mathbf{x}=\langle \mathbf{S}, \mathbf{A} \rangle$, with visual and proprioceptive states $\mathbf{S}=\{\mathbf{s}^t\}_{t=1}^T$ and applied actions $\mathbf{A}=\{\mathbf{a}^t\}_{t=1}^T$, we use cross-correlation summary statistics to compute the MDNN input, as:
\begin{equation}
    \psi(\mathbf{S}, \mathbf{A})=(\{\langle \mathbf{S}_i, \mathbf{A}_j \rangle\}_{i=1,j=1}^{D_s,D_a}, \mathrm{E}[\mathbf{S}],\mathrm{Var}[\mathbf{S}]),
\end{equation}
where $D_s$ is the state space dim., $D_a$ is the action space dim., $\langle \cdot,\cdot \rangle$ denotes the dot product, $\mathrm{E}[\cdot]$ is the expectation (mean) and $\mathrm{Var}[\cdot]$ the variance (std dev)~\cite{ramos2019bayessim}. For readability throughout the text, we interchangeably denote the raw $\mathbf{x}$ and summarised $\psi(\mathbf{S},\mathbf{A})$ trajectories, unless otherwise required.

\subsection{Reward design}
\label{subsec:reward-design}

The DLO keypoints are temporally consistent, but permutation invariance is not guaranteed~\cite{antonova2022bayesian}, so we cannot track specific DLO parts.
Not having an explicit tip~\cite{lim2022real2sim2real, chi2024iterative} to guide hinders learning a generalist policy, but raises the practicality of specialist policies for specific sets of DLO parameterisations through a Real2Sim2Real treatment. We condition policies for whole-body guidance with a reward function that evaluates only the rollout outcome, i.e., which cubes were hit. 

In each state $s_t$, we assume a base reward $r_t=0.0$ and check whether the \emph{top cube} (\setulcolor{yellow}\ul{yellow}) has been hit, in which case we assign a reward $r_t=2.0$. We then inspect the stack in reverse order (top to bottom, excluding the top for which we have just checked), and for every hit cube with index $i$ we apply an exponential reward decay, as $r_t \mathrel{-}= 0.05 \times 2^{4-i}$. Thus, for $[$\setulcolor{red}\ul{red}, \setulcolor{green}\ul{green}, \setulcolor{blue}\ul{blue}, \setulcolor{orange}\ul{orange}, \setulcolor{purple}\ul{purple}$]$ (excl. top \setulcolor{yellow}\ul{yellow}) cube indexes iterated in reverse as $[$\setulcolor{purple}\ul{4}, \setulcolor{orange}\ul{3}, \setulcolor{blue}\ul{2}, \setulcolor{green}\ul{1}, \setulcolor{red}\ul{0}$]$ we can have the respective penalties $[$\setulcolor{purple}\ul{$0.05$}, \setulcolor{orange}\ul{$0.1$}, \setulcolor{blue}\ul{$0.2$}, \setulcolor{green}\ul{$0.4$}, \setulcolor{red}\ul{$0.8$}$]$ applied in $r_t$. 

Thus, for simplicity within the scope of this paper, 
the penalty for a cube is applied regardless of whether the cubes above it have been hit. 
However, we acknowledge that a stricter reward formulation, such as explicitly conditioning penalties on strike order or on verified interactions, could be more robust to artefacts such as missed cube detections.

A cube is \emph{hit}/\emph{whipped}/\emph{knocked} if its keypoint's pixel-space Euclidean distance (L2 norm) of the current and initial positions is $>0.06$. 
The episode ends when the top cube is hit. 

\subsection{Domain randomisation, training \& Sim2Real deployment}
\label{subsec:policy-learn-n-deploy-setup}

We train \emph{proximal policy optimisation} (PPO) agents~\cite{schulman2017proximal} in \emph{vectorised} environments using Stable Baselines3~\cite{stable-baselines3}. 
DR is performed across $12$ parallel environments, each sampling a distinct $\boldsymbol{\theta}$ from a domain distribution $p(\boldsymbol{\theta})$~\cite{kamaras2025distributional}. 
This limited sample budget reflects the cost of deformable object simulation and increases the importance of a descriptive $p(\boldsymbol{\theta})$.
MoG domain distributions are sampled with \emph{low-variance} to ensure coverage of all components. 
For fixed support, we sample over $\Theta_0$; for adapted support, over $\Phi$, with outliers clipped as in \Cref{subsec:lv-exper}. 
To improve Sim2Real robustness, we additionally randomise camera and stack placement: camera positions are uniformly perturbed by $(\pm2.5,\,\pm2.5,\,\pm2.5)\,\si{\centi\meter}$ in $(x,y,z)$, and stack positions by $\pm2\,\si{\centi\meter}$ in $x$ and $\pm5\,\si{\milli\meter}$ in $y$. 

We train PPO with default hyperparameters~\cite{schulman2017proximal} for $60,000$ steps, using an episode length of $16$ and a batch size of $16$. 
For real-world deployment, we extend a sample-efficient robotic RL framework~\cite{luo2024serl} and empirically tune the impedance controller so that $16$ real steps match $16$ simulation steps~\cite{kamaras2025distributional}.

\section{Real2Sim2Real DLO whipping results}
\label{sec:dlo-results}

We address the following questions:
\begin{enumerate}
    \item Can BayesSim-EDGE infer a \textbf{higher fidelity} posterior and \textbf{adapt the posterior's support} for \textbf{more \emph{realistic} simulation} in a visuomotor DLO whipping task?
    \item What is the impact of posterior-driven support adaptation in Real2Sim2Real for this task? We \textbf{cross-evaluate} over a parametric set of (real) DLOs for: \begin{enumerate}
        \item \textbf{Classification granularity} of different DLO $\boldsymbol{\theta}$.
        \item \textbf{Object-centric agent adaptation}.
        \item \textbf{Perceived challenge} of each DLO to our agents.
    \end{enumerate}
\end{enumerate}

\begin{figure}[!t]
    \centering
    \includegraphics[width=1.0\columnwidth]{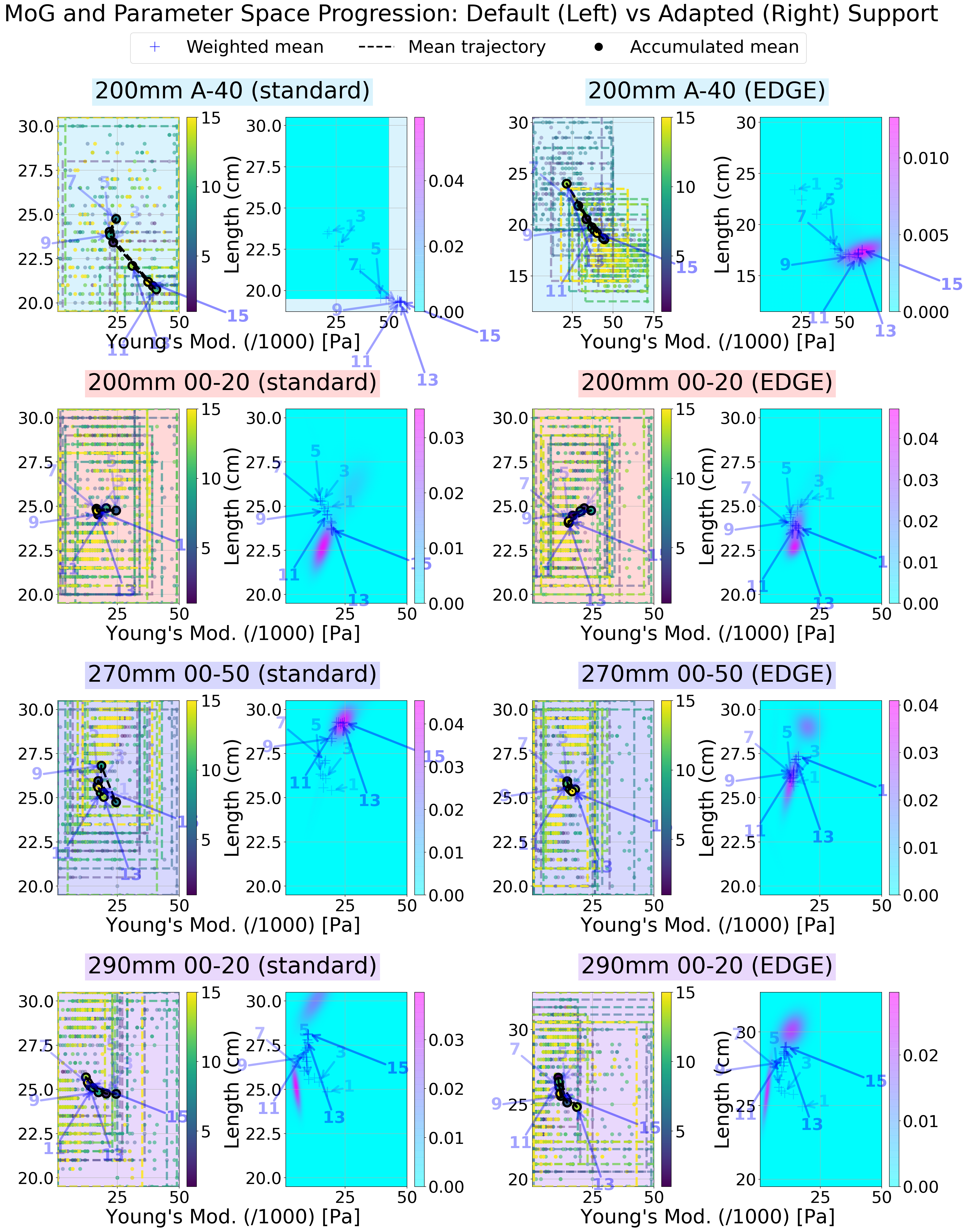}
    \caption{Prior samples' progression (left) and final MoG posteriors (right) after $15$ inference iterations for DLOs-\{\plotrefblue{0}, \plotrefred{1}, \plotrefpurple{2}, \plotrefviolet{3}\}.
    Scatterplots show the per-iteration sample progression and their bounding boxes. Colours indicate which iteration samples were drawn. 
    The bigger dots mark the accumulated dataset's mean every second iteration. 
    On MoG plots, component means are marked in blue crosses and colours show likelihood. MoG progression is annotated with arrows pointing to the position of every second iteration's weighted mean. 
    }
    \label{fig:mog-n-param-space-progr}
\end{figure}

\begin{figure}[!t]
    \centering
    \includegraphics[width=1.0\columnwidth]{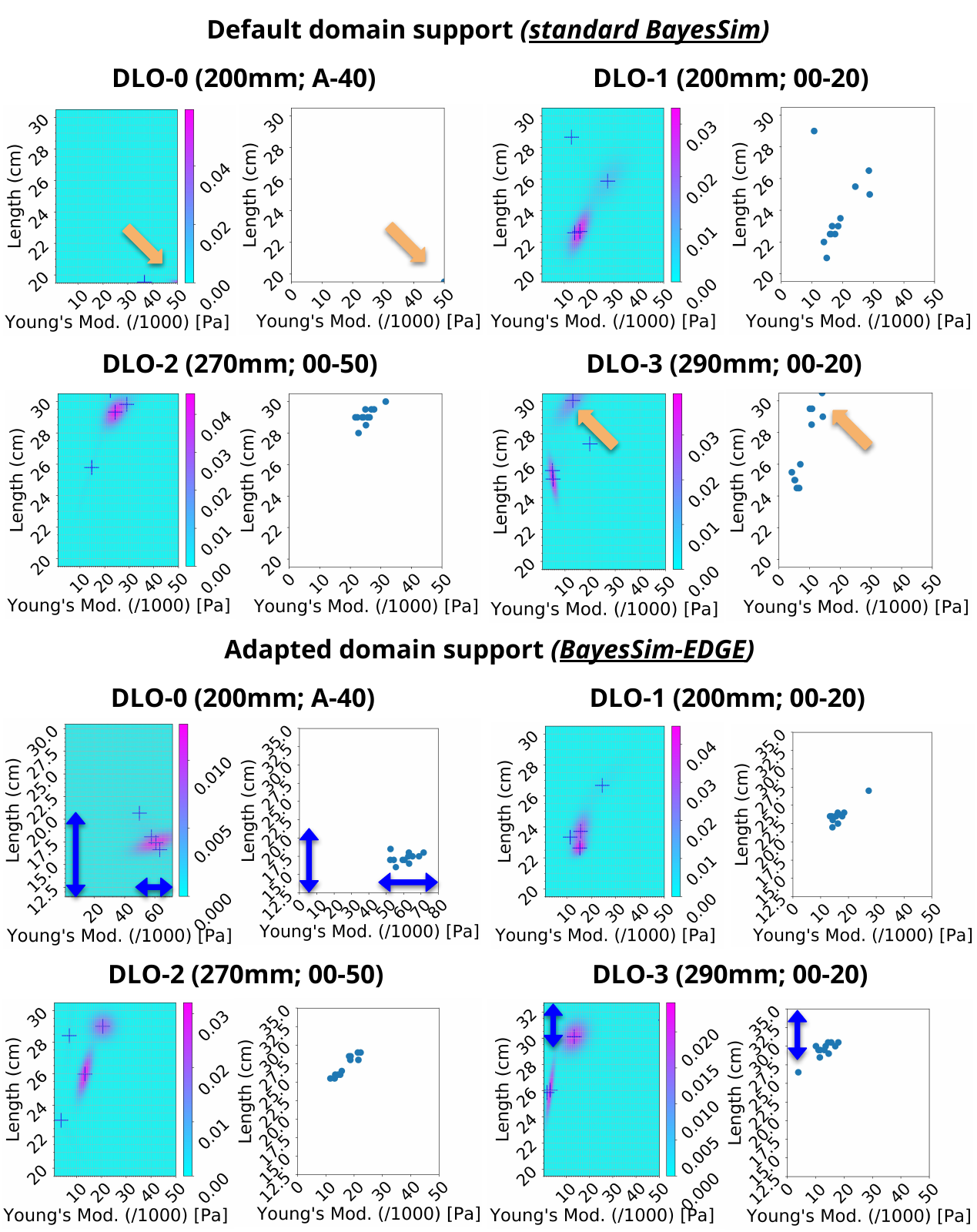}
    \caption{Inferred MoG posteriors after $15$ iterations and the domain samples drawn when each MoG is used for DR. Component means are shown in blue crosses and colourbars quantify likelihood. On the top, we see the results of BayesSim sampling over the default, potentially misspecified, support. Orange arrows indicate the need for adaptation (DLOs-\{0, 3\}). On the bottom, we see the results of BayesSim-EDGE iteratively adapting the support (blue arrows).}
    \label{fig:mog-posterior-for-dr}
\end{figure}

\subsection{Support adaptation impact on LFI}
\label{subsec:real2sim-res--lfi}

\Cref{fig:mog-n-param-space-progr} shows the 2D MoG posteriors from Real2Sim inference alongside the domain samples at each iteration (\cref{alg:bsim}, \cref{alg:bsim:line:dataset-update}). 
Sample density increases in regions of high posterior belief over $\boldsymbol{\theta}$. 
Each posterior is a $4$-component MoG, whose means represent \emph{alternative hypotheses} over a DLO's parameters. 
Overall, \textbf{Young's modulus} (softness) \textbf{is inferred} \textbf{more precisely than length}, which remains the dominant source of uncertainty when tracking a DLO with $4$ keypoints, as reflected by component variance and mean dispersion.

\Cref{fig:mog-posterior-for-dr} shows the final MoGs, highlighting cases of support misspecification and the corresponding adaptations.
For the shorter and stiffer DLO-0 the standard BayesSim places both length and softness at the boundaries of $\Theta_0$, indicating misspecified support; 
the \textbf{EDGE} heuristic 
\textbf{expands the support toward more likely parameterisations}. 
For the longer and softer DLO-3, standard BayesSim confidently infers softness within $\Theta_0$ but suggests length lies beyond its upper bound, 
which EDGE expands to improve confidence.

For longer DLOs-$\{2, 3\}$, alternative hypotheses primarily vary along the length dimension. This variance appears in both standard BayesSim and EDGE for DLO-3, but only in EDGE for DLO-2, reflecting the stochasticity in Real2Sim LFI for dynamic DLO control with a fixed number of keypoints.

\begin{figure}[!t]
    \centering
    \includegraphics[width=1.0\columnwidth]{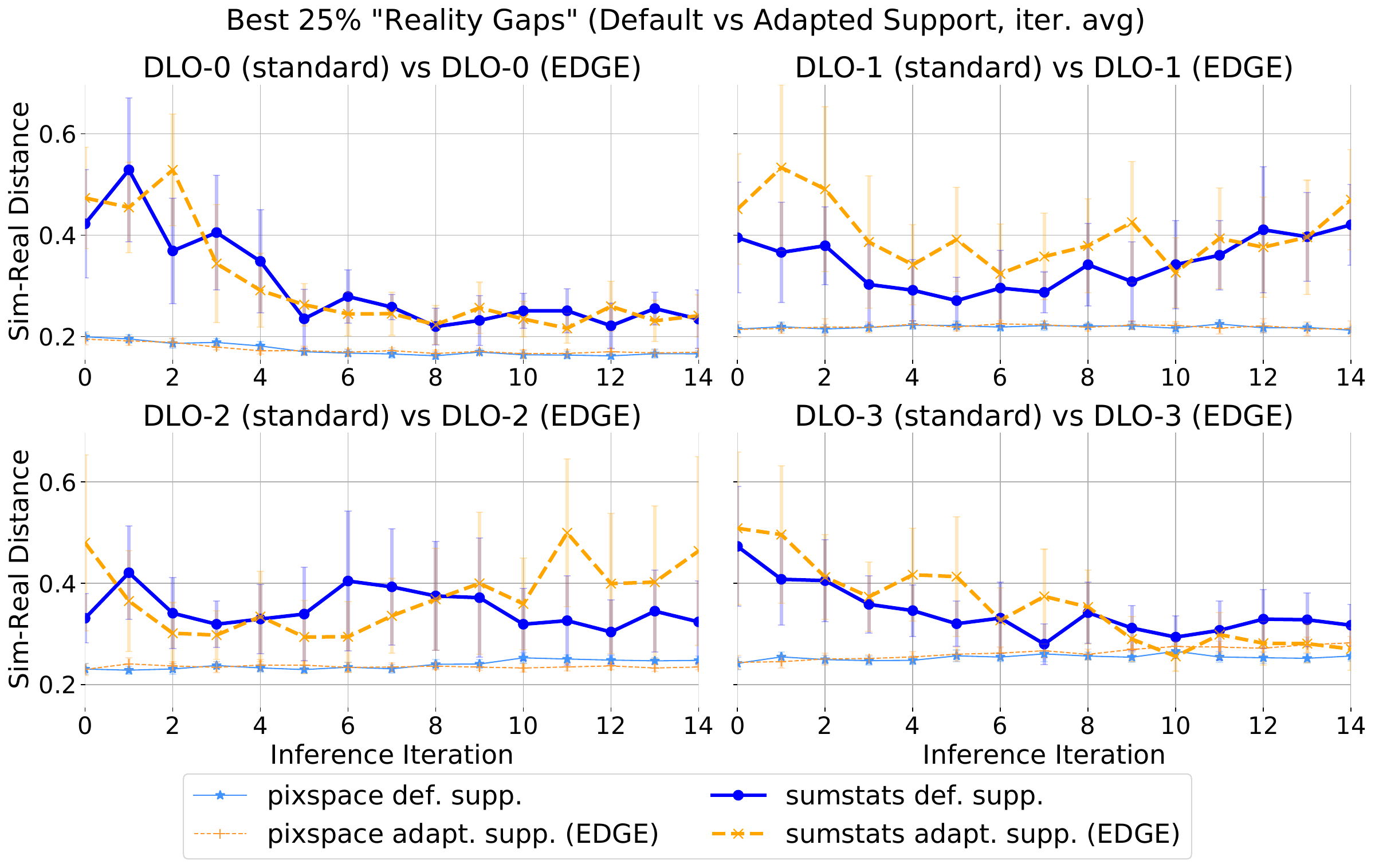}
    \caption{Pixel-space (faded) and statistical (bold) reality gap of the $15$ inference iterations with default and adapted support of \cref{fig:mog-n-param-space-progr}. 
    We compute the bidirectional Chamfer distances of the averaged timesteps of the simulated and real keypoint trajectories (pixel-space) and the Euclidean distances of the cross-correlation summary statistics. 
    Errorbars show standard deviations. We average and report the distances of each iteration's $25\%$ best-scoring samples. 
    }
    \label{fig:sim2real-distance-progr}
\end{figure}

\Cref{fig:sim2real-distance-progr} reports the reality gap of prior samples across inference iterations (\cref{fig:mog-n-param-space-progr}), 
measured in both pixel space and summary statistics space.
For both fixed (default) and adapted support, the largest convergence occurs for the stiffer DLO-0 and the longer and softer DLO-3, which deviate most from domain median $\mu$. 
Extrinsic differences between standard BayesSim and EDGE are small under these metrics. 
Notably, the summary-statistics gap is more informative than the pixel-space gap, highlighting the importance of \textbf{evaluating inference performance in the MDNN input space} rather than in task-level features.

\subsection{Uncertainty over parameter estimation is reflected in DR}
\label{subsec:real2sim-res--dr}

\Cref{fig:mog-posterior-for-dr} shows each MoG's $12$ domain samples, which are used for object-centric policy training. 
We see how \textbf{the posterior certainty} (\emph{sharpness}) in a $\boldsymbol{\theta}$ dimension 
\textbf{is reflected in the clustering of domain samples}. 
The spread of the samples also shows that some of the alternative hypotheses for $\boldsymbol{\theta}$ contribute to the low-variance sampling. 
This is most evident in the softer DLOs-$\{1, 3\}$ results for both standard BayesSim and EDGE, and in the DLO-0 result of EDGE. 
For EDGE, we also see how the low-variance sampling balances alternative hypotheses for DLOs-$\{2, 3\}$ with in-between clusters of samples. 
In general, \textbf{adapted support experiments result in cleaner clusters of domain samples}, which shows that \textbf{LFI benefits from a reasonably broader support}.

\subsection{Real2Sim2Real for object-centric agent adaptation}
\label{subsec:policy-learn-sim2real-res}

\begin{figure}[!t]
    \centering
    \includegraphics[width=0.85\columnwidth]{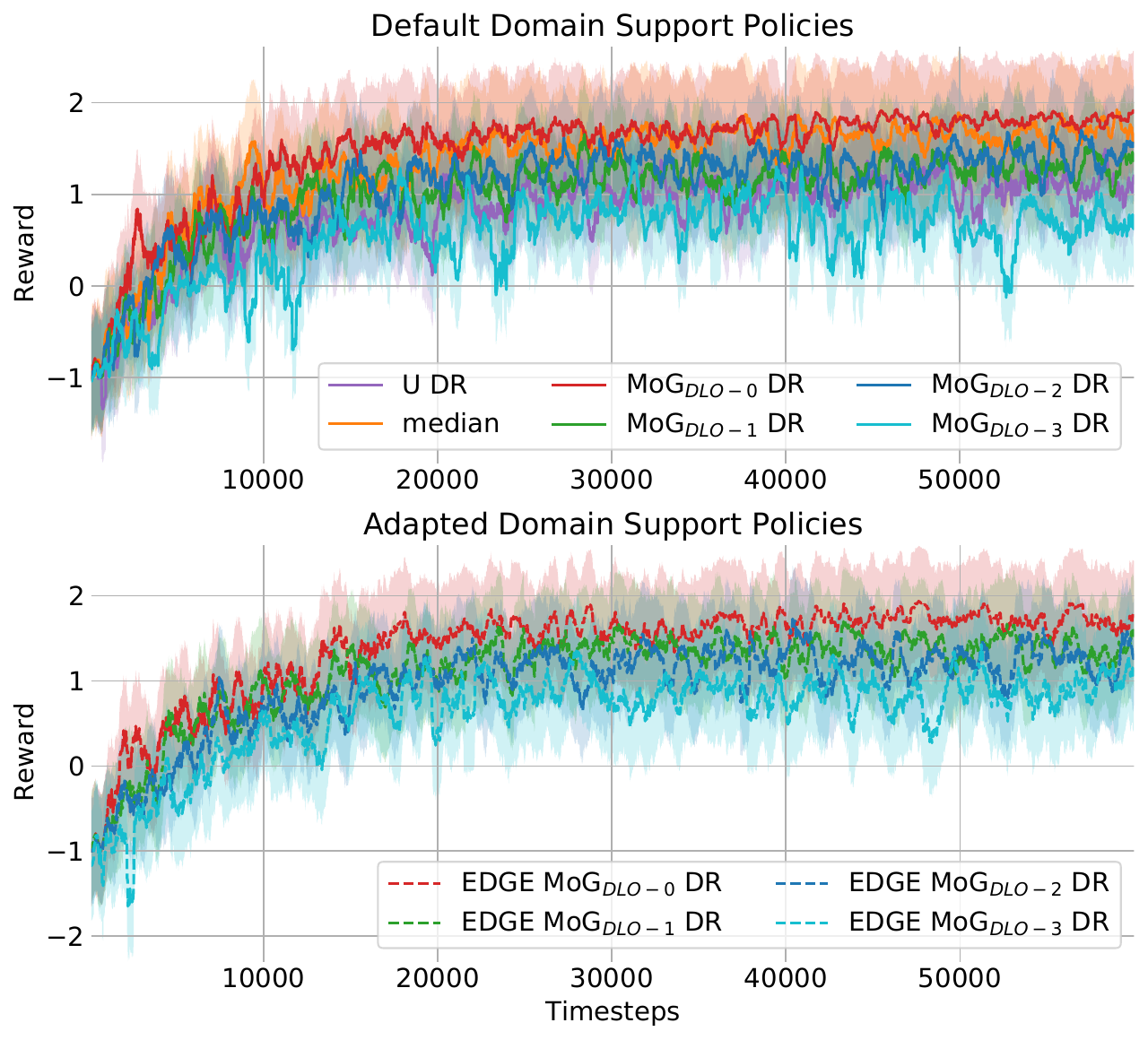}
    \caption{Learning curves of PPO agent training when performing DR using different domain distributions, producing the \cref{fig:mog-posterior-for-dr} domain samples. On the top, we have the agents using the default, potentially misspecified, support. On the bottom, we have the agents sampling from domain distributions with a support that has been heuristically adapted using BayesSim-EDGE.}
    \label{fig:learning-curves}
\end{figure}

We train four PPO agents using the MoG posteriors from the default support $\Theta_0$ and four using EDGE-adapted posteriors, along with a uniform DR agent on $\Theta_0$ (\setulcolor{purple}\ul{$\text{PPO-}\mathit{U}$}) and a median-parameter agent (\setulcolor{orange}\ul{$\text{PPO-}\mu$}).  
All policies are underlined per \Cref{fig:learning-curves}. 
We cross-evaluate the $10$ policies on $4$ real DLOs and a simulated median ($\mu$) DLO, with $4$ repetitions per setting, yielding $160$ Sim2Real and $40$ Sim2Sim evaluations. 

\begin{figure*}[!t]
    \centering
    \includegraphics[width=0.95\textwidth]{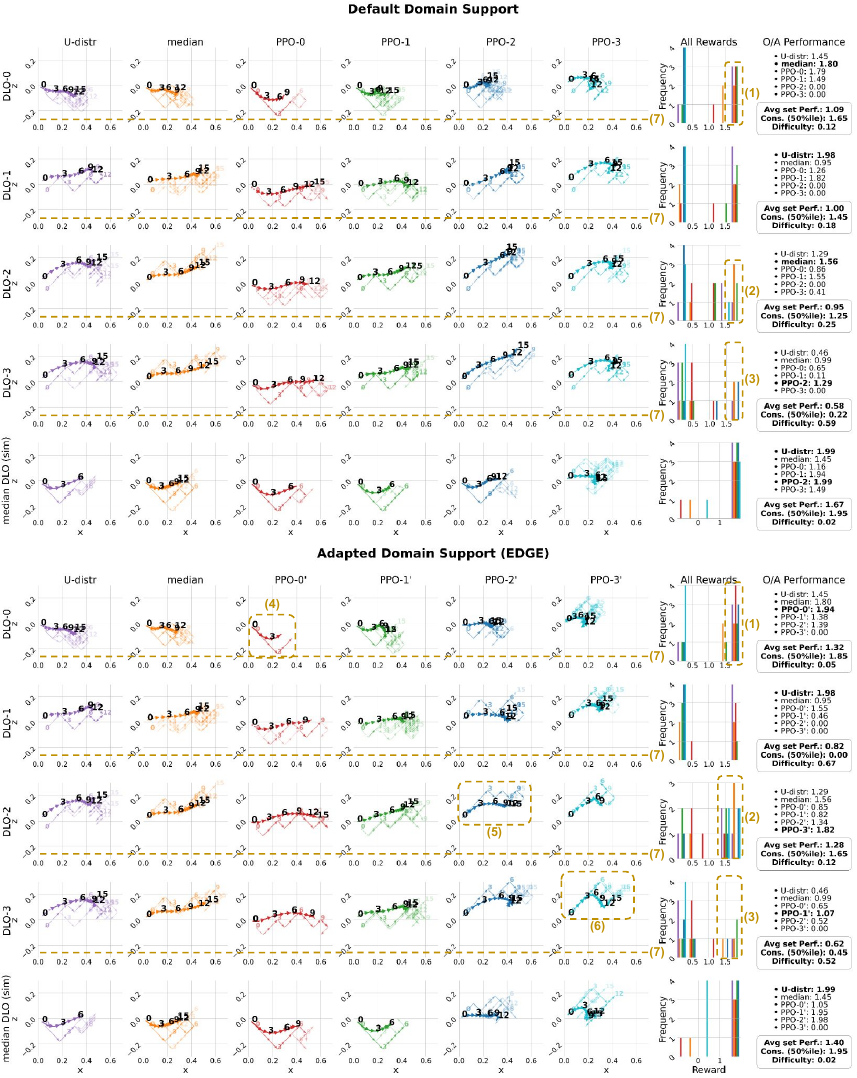}
    \caption{
    EEF trajectories resulting from deploying in real world (rows 1-4 \& 6-9; $4$ real DLOs) and in simulation (rows 5 \& 10; median DLO) the policies trained while using for DR the posteriors inferred over the \emph{default} (top) and \emph{adapted} (bottom) \emph{support} (columns 1-6), illustrated in \cref{fig:mog-posterior-for-dr}. 
    We repeat each deployment $4$ times and plot each episode's measured accumulation of commanded EEF translations along the $x$ and $z$ axes 
    (commanded \emph{actions}, $\langle dx, dz \rangle$) 
    along with the respective soft-DTW barycentre (bolder lines).
    To indicate trajectories' duration, we annotate every three timesteps from their beginning.
    Although $\text{PPO-}\mathit{U}$ and $\text{PPO-}\mu$ are default support policies, we plot them on both top and bottom grids for visual consistency. 
    The histograms (col. 7) show all collected episode rewards on each DLO (row), 
    with bin colours matching trajectory line colours,
    and the last column (col. 8) gives a performance overview of the set of agents for each row's DLO.
    }
    \label{fig:def-adapt-supp-eef-trajs-plots}
\end{figure*}

\Cref{fig:def-adapt-supp-eef-trajs-plots} shows the accumulated EEF $\langle dx, dz \rangle$ trajectories over $4$ repetitions and their soft dynamic time warping (soft-DTW) barycentres, which cluster and average inhomogeneous timeseries~\cite{cuturi2017soft}. 
It also reports: 
\begin{enumerate*}[label={(\roman*)}]
    \item per-agent average reward (columns); 
    \item average reward per DLO across agents (Avg set Perf.); 
    \item the median reward across agents as a consistency metric (Cons.); 
    \item DLO difficulty, defined as $1-(\text{Cons.}-r_{\min})/(r_{\max}-r_{\min})$, given global reward extrema $r_{\min},r_{\max}$, with higher values indicating lower robustness. 
\end{enumerate*} 

\begin{figure}[!t]
    \centering
    \includegraphics[width=1.0\columnwidth]{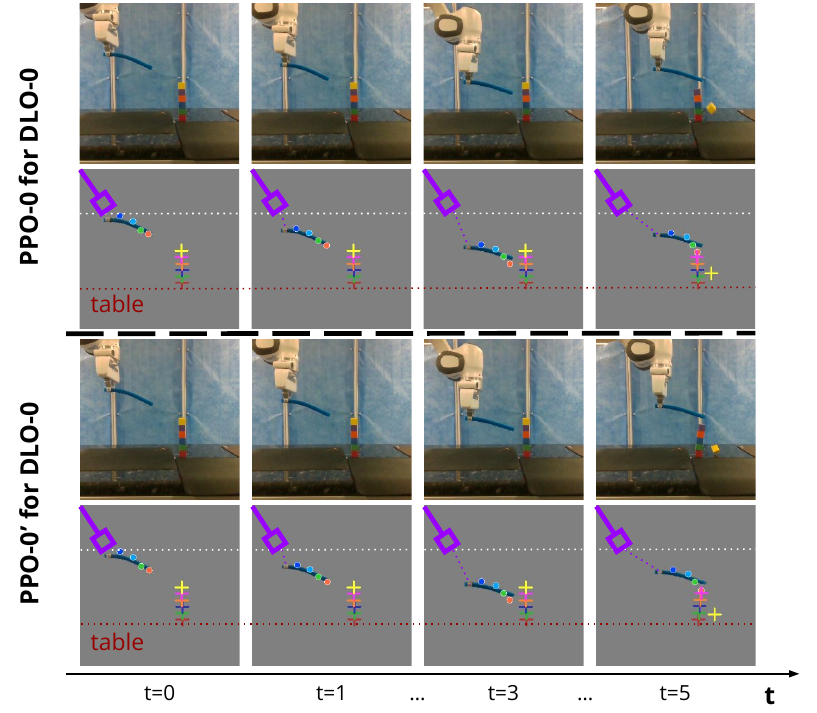}
    \caption{
    Extrinsic evaluation timelapses of real-world PPO-$0$ and $0'$ policy evaluations using DLO-0, in conjunction with \cref{fig:def-adapt-supp-eef-trajs-plots}.
    A best-case outcome can result in almost identical trajectories, despite PPO-$0$ being generally less stable.
    Dotted white and red lines mark the initial EEF $z$ and table surface respectively. A purple indicator on top left of each image marks the initial EEF position, with dotted purple lines showing its displacement.
    }
    \label{fig:eef-trajs-timelapse-dlo-0}
\end{figure}

\begin{figure*}[!t]
    \centering
    \includegraphics[width=1.0\textwidth]{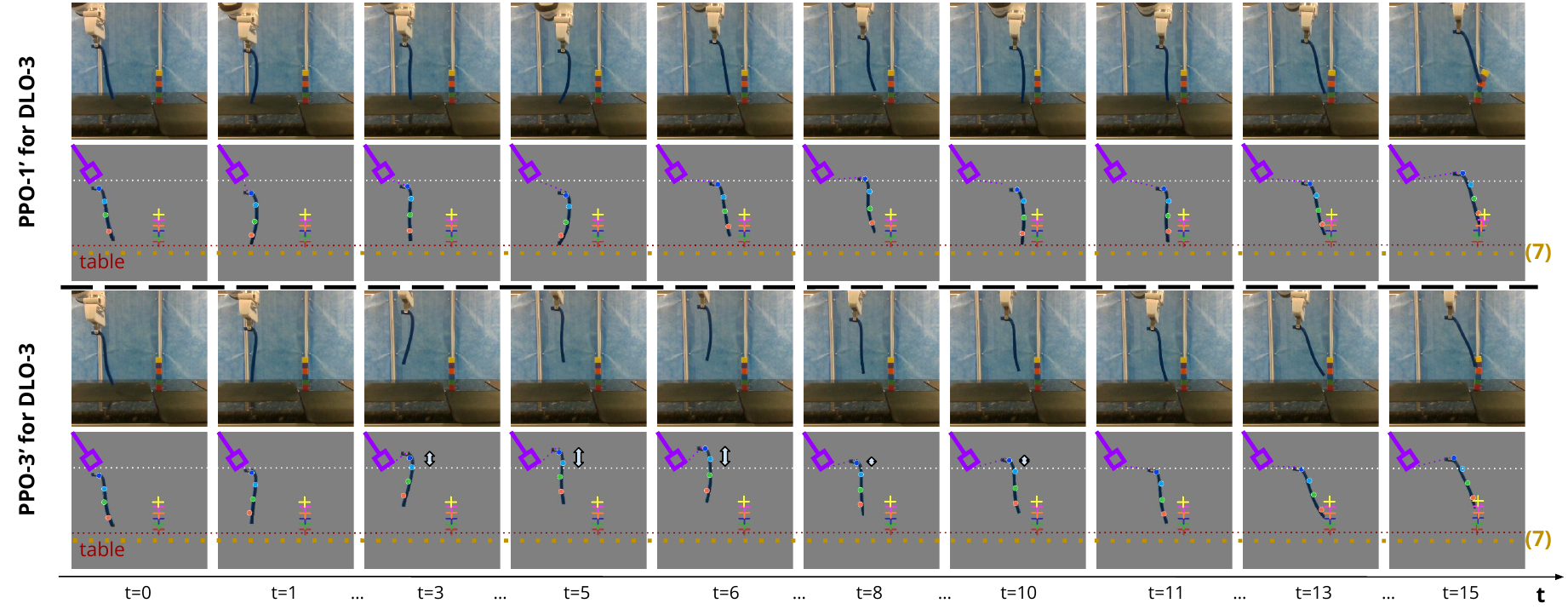}
    \caption{
    Extrinsic evaluation timelapses of real-world PPO-$1'$ and $3'$ policy evaluations using DLO-3, in conjunction with \cref{fig:def-adapt-supp-eef-trajs-plots}.
    Annotations follow \cref{fig:eef-trajs-timelapse-dlo-0}.
    We observe the greater DLO distance from the table on $t \in [3, 10]$, annotated with light blue arrows with black border along the PPO-$3'$ trajectory.
    }
    \label{fig:eef-trajs-timelapse-dlo-3}
\end{figure*}

We further observe EEF motion patterns (highlighted in \textcolor{olive}{\textbf{olive}}) that reflect the domain distributions used during training~\cite{kamaras2025distributional}, indicating adaptation of agent behaviour to the inferred physical parameters of each DLO. 
\Cref{fig:eef-trajs-timelapse-dlo-0,,fig:eef-trajs-timelapse-dlo-3} show representative examples.

Reward histograms show that DR with adapted support yields higher returns, especially for $\text{EDGE-MoG}_{\text{DLO-0}}$ (\dashulinecolor{$\text{PPO-}0'$}{black}{red}) over $\text{MoG}_{\text{DLO-0}}$ (\setulcolor{red}\ul{$\text{PPO-}0$}) \textcolor{olive}{\textbf{(1)}} and $\text{EDGE-MoG}_{\text{DLO-2}}$ (\dashulinecolor{$\text{PPO-}2'$}{black}{blue}) over $\text{MoG}_{\text{DLO-2}}$ (\setulcolor{blue}\ul{$\text{PPO-}2$}) \textcolor{olive}{\textbf{(2)}}. 
In contrast, the very short and soft DLO-1 is well covered by the default support $\Theta_0$ and therefore $\text{MoG}_{\text{DLO-1}}$ (\setulcolor{green}\ul{$\text{PPO-}1$}) is more robust than $\text{EDGE-MoG}_{\text{DLO-1}}$ (\dashulinecolor{$\text{PPO-}1'$}{black}{green}) due to being trained over a better spread domain distribution (\cref{fig:mog-posterior-for-dr}).

For the longer and softer DLO-3, support adaptation improves Sim2Real performance when $\text{EDGE-MoG}_{\text{DLO-3}}$ (\dashulinecolor{$\text{PPO-}3'$}{black}{cyan}) controls DLO-2. 
When controlling DLO-3, PPO-$3'$ produces useful but insufficiently fast motions to fully affect the top cubes within 16 steps (\cref{fig:eef-trajs-timelapse-dlo-3}, bottom). 
In contrast, $\text{PPO-}1'$, trained on the similarly soft but shorter DLO-1, attains the optimal reward in $2/4$ rollouts \textcolor{olive}{\textbf{(3)}}, although extrinsic evaluation finds its behaviour useful in only $1/4$ cases. 
Finally, while $\text{MoG}_{\text{DLO-3}}$ (\setulcolor{cyan}\ul{$\text{PPO-}3$}) transfers partially to DLO-1, the over-adapted $\text{PPO-}3'$ does not, reflecting its specialisation to longer bodies.
This mismatch between \emph{expected} and \emph{actual} $\text{PPO-}3'$ performance highlights a reality gap, showing how 
\textbf{an agent's skill with real DLOs is shaped by the sim domain that is derived from a \emph{perceived} parameterisation}.

Overall, large support adaptation in DLO-$0$ inference (\cref{fig:mog-posterior-for-dr}) yields a corresponding performance gain for PPO-$0'$. 
DLO-$1$ requires no adaptation, but the sharper BayesSim-EDGE posterior results in \emph{median performance} for PPO-$1'$. 
For DLO-$2$, the EDGE-inferred MoG produces PPO-$2'$, which is more effective than PPO-$2$. 
Finally, although EDGE adapts the length for DLO-$3$, 
due to the inferred stiffness and the assumed density PPO-$3'$ is most useful for DLO-$2$.

Sim2Sim results on DLO-$\mu$ confirm the \emph{median performance} trend, with policies trained via DR on DLOs-$\{1, 2\}$ showing the greatest robustness, along with $\text{PPO-}\mathit{U}$. This aligns with the proximity of DLOs-$\{1,2\}$ to the parameter-space median $\mu$ (\cref{subsec:real2sim-res--lfi}).

Extrinsic evaluation shows that \textbf{adapted-support policies} PPO-$\{0', 2', 3'\}$ \textbf{perform more stably on their corresponding DLOs}-$\{0,2,3\}$ \textcolor{olive}{\textbf{(4-6)}}, e.g., via shorter soft-DTW trajectories for DLO-$0$. 
In contrast, PPO-$1'$ exhibits higher rollout variance and lower performance on DLO-$1$ than PPO-$1$, but remains useful on longer DLOs-$\{2, 3\}$, performing comparably to PPO-$2'$. 
Overall, PPO-$\{0', 1', 2', 3'\}$ adapt better to DLO length than PPO-$\{0, 1, 2, 3\}$, as reflected by EEF height relative to the table \textcolor{olive}{\textbf{(7)}}. 
This reduces the risk of dragging the DLO and shows that \textbf{appropriate DR distributions and support} can mitigate such interactions \textbf{without complicating the RL formulation}.



\section{Conclusion}
\label{sec:conclusion}

LFI typically assumes an arbitrary, fixed support, which can lead to misspecification and propagate errors in Real2Sim2Real pipelines. 
We show that posterior signals---such as probability mass accumulation and mode shifts toward domain boundaries---can reveal misspecified support and guide its adaptation alongside posterior inference. 

We address this empirically with three posterior-driven BayesSim heuristics (EDGE, MODE, and CENTRE). 
Sim2Sim experiments on the Lotka-Volterra model show that EDGE is the most robust and easiest to tune. 
Applied to Real2Sim2Real visuomotor DLO whipping, EDGE improves BayesSim's ability to infer fine-grained parameter differences and increases object-centric adaptation and task performance on $3/4$ real DLOs (idx. $\{0, 2, 3\}$). 

Posterior-driven support adaptation with deterministic rules may be redundant in some settings, as LFI can over-attribute evidence to nearby samples within the effective support. 
While careful tuning mitigates this effect, future work could replace explicit heuristics with automated rule extraction~\cite{brotherton1999automated, murdoch2017iclr}.


\appendices 

\section{Misspecified Support impact in Sequential Neural Posterior Estimation}
\label{appendix:snpe-misspec-support}

Sequential neural posterior estimation~\cite{papamakarios2019sequential} comprises formally robust posterior inference methods. While BayesSim is preferred in this paper for its practical merits in robotics, as discussed in~\Cref{subsec:lfi-prelim}, we additionally evaluate Automatic Posterior Transformation (APT)~\cite{greenberg2019icml} to assess whether the highlighted support misspecification issue extends to the broader landscape of SBI methods. We follow the implementation in~\cite{boelts2025jossw}, which treats APT as the reference normalising flow method~\cite{papamakarios2021normalizing} for neural posterior estimation.

We experiment with Lotka-Volterra, using as observation $\mathbf{x}$ the same summary statistics as in \cref{subsec:lv-exper}, but this time simulating the system as a nonlinear ODE with the addition of independent Gaussian noise, instead of an MJP, to also experimentally cover the approach of~\cite{greenberg2019icml, boelts2025jossw}.

We consider $\boldsymbol{\theta}_{\text{gt}} = [0.01, 0.1, 0.1, 0.02]$, default (\plotrefblue{`OK'}) support \plotrefblue{$[[0.005, 0.03],$ $[0.005, 0.15],$ $[0.05, 0.15],$ $[0.01, 0.03]]$} and \plotrefred{misspecified} support \plotrefred{$[[0.015,$ $0.04],$ $[0.12, 0.3],$ $[0.05, 0.15],$ $[0.01, 0.03]]$}. Following~\cite{boelts2025jossw}, we experiment with initial populations $(X, Y) = (9, 40)$, simulating $200$ steps with $dt=0.1$.

\begin{figure}[t]
\centering
\includegraphics[width=1.0\linewidth]{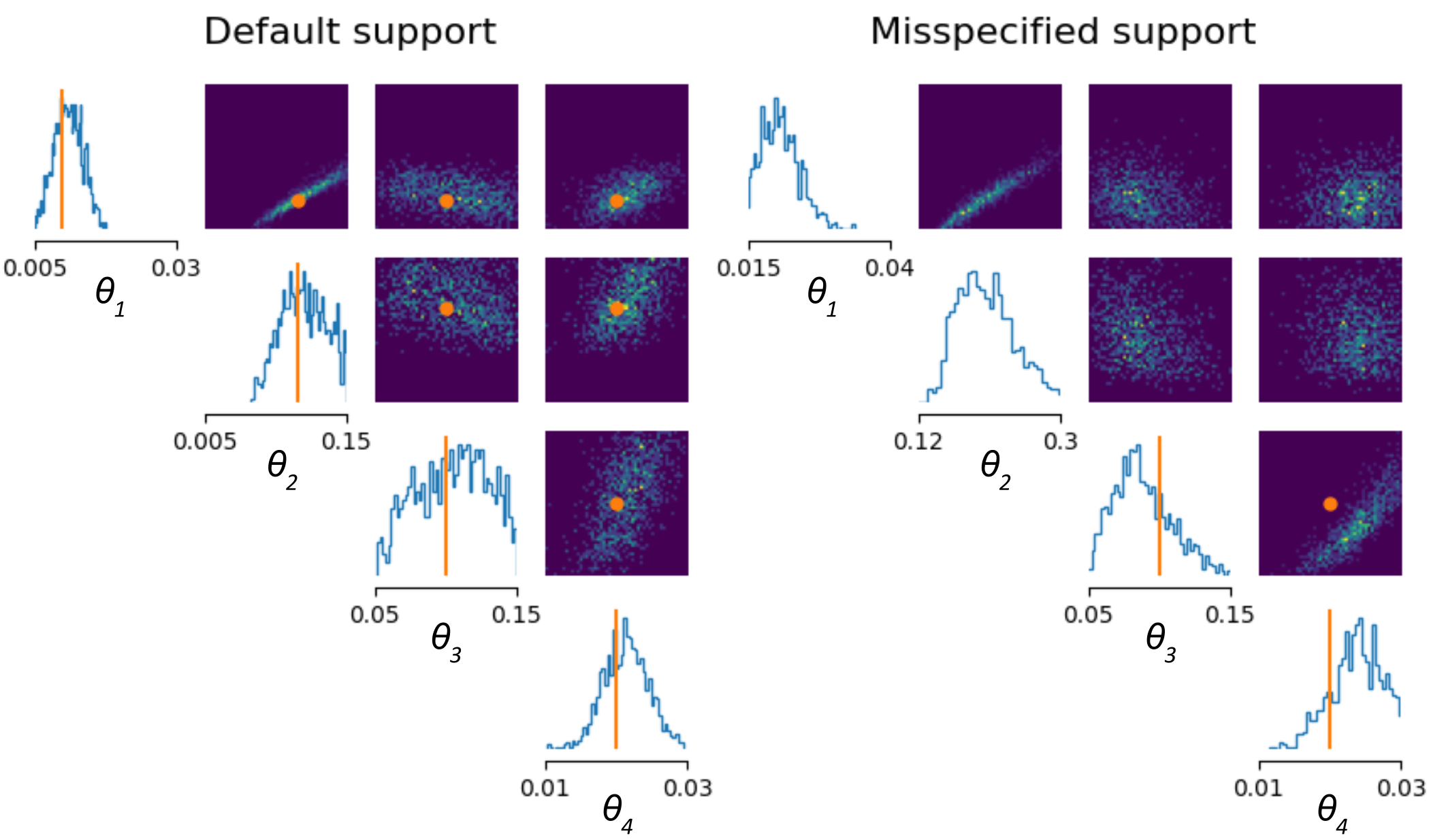}
\includegraphics[width=0.49\linewidth]{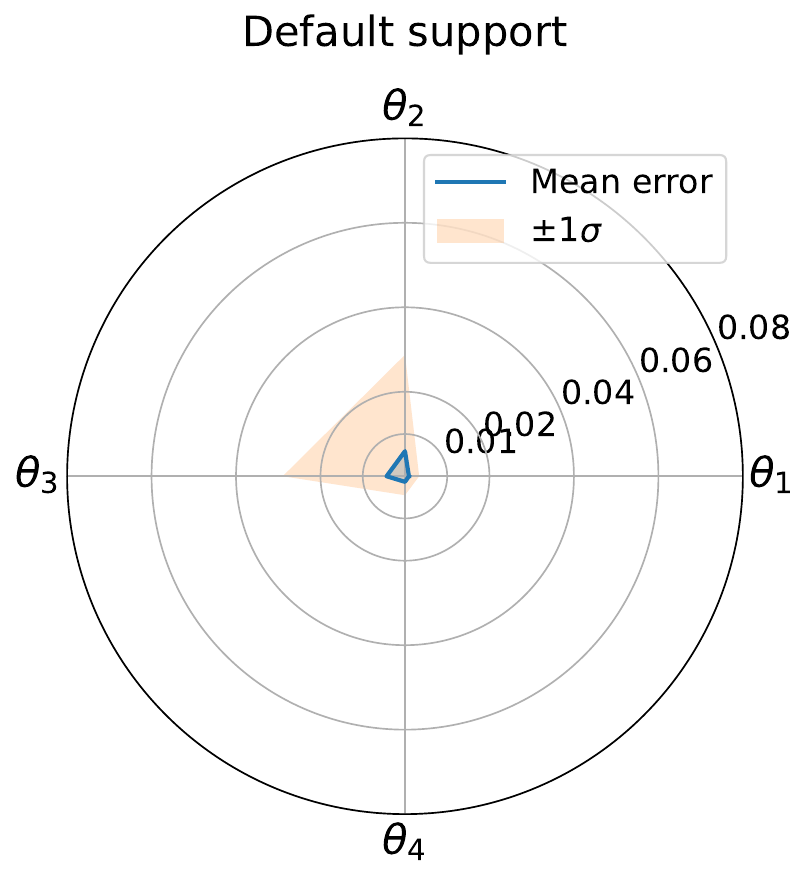}
\includegraphics[width=0.49\linewidth]{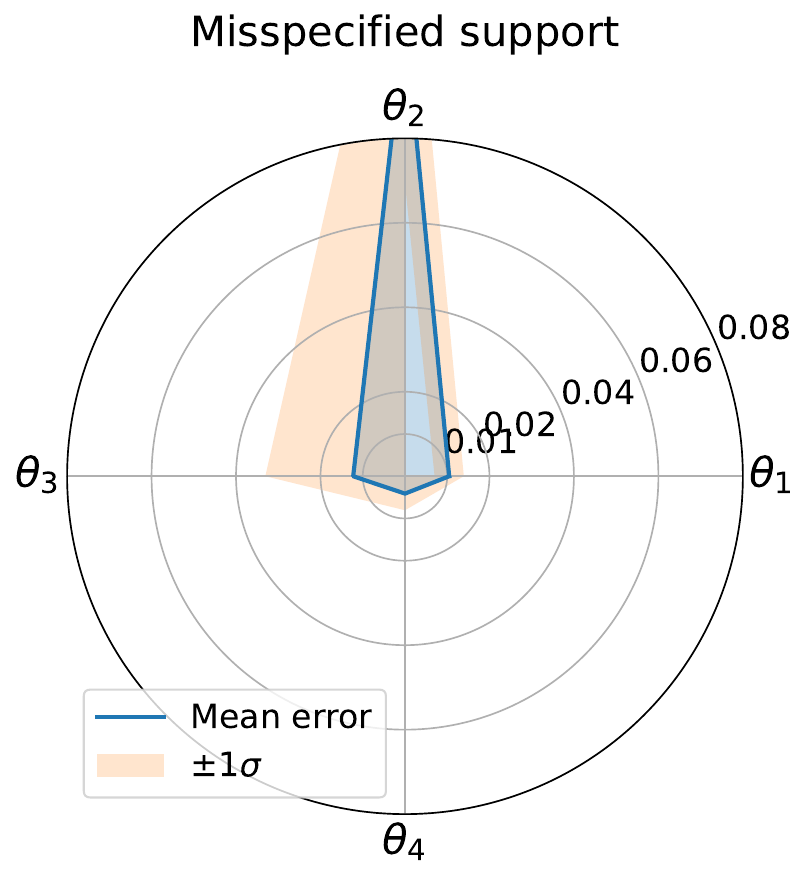}
\includegraphics[width=0.49\linewidth]{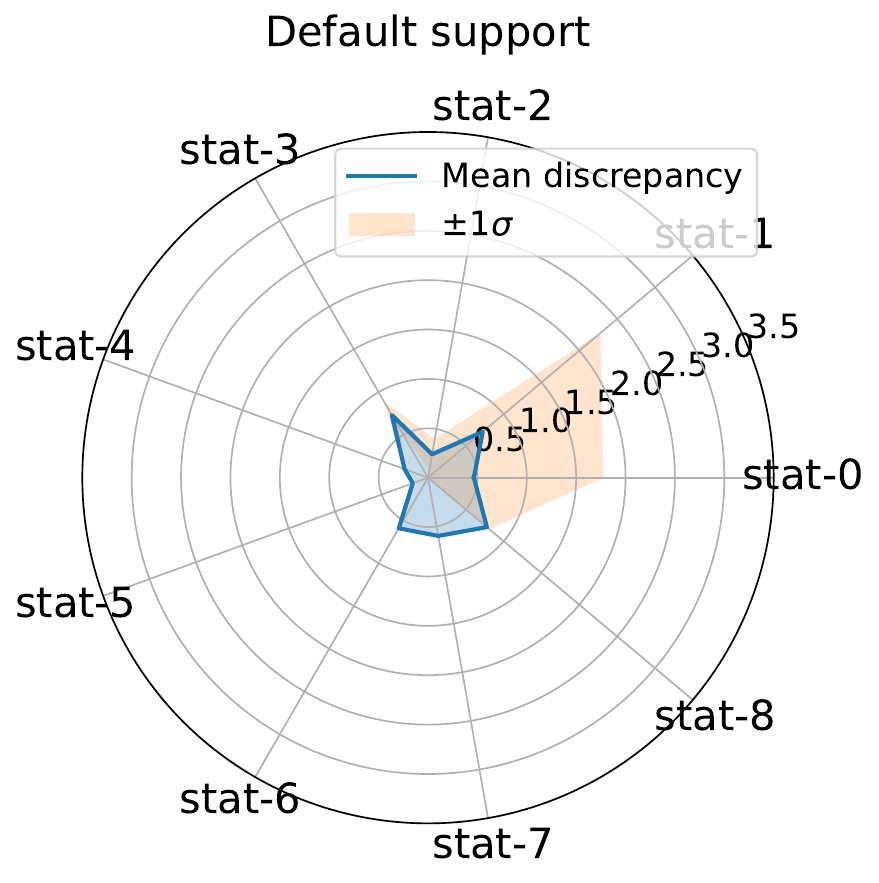}
\includegraphics[width=0.49\linewidth]{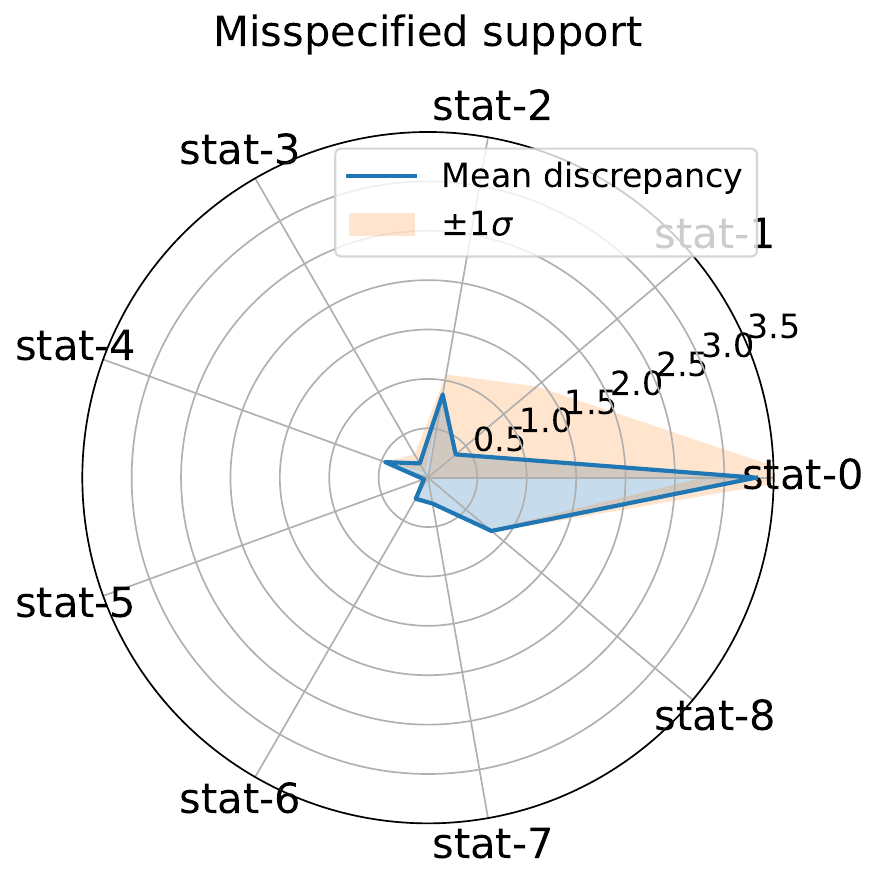}
\caption{
\emph{Top row:} per-parameter-pair posteriors from ATP inference over the default (left) and misspecified support (right), as defined in~\cref{appendix:snpe-misspec-support}.
\emph{Middle row:} the mean discrepancy and standard deviation ($\sigma$) of $10$ posterior samples of $\boldsymbol{\theta}$ relative to $\boldsymbol{\theta}_\text{gt}$, for inference over the default (`OK') support (left) and the misspecified support (right).
\emph{Bottom row:} the mean discrepancy and standard deviation of the summary statistics computed from the trajectories corresponding to the $10$ posterior samples of $\boldsymbol{\theta}$ and to $\boldsymbol{\theta}_\text{gt}$.
}
\label{fig:lv-atp-posteriors-and-polars}
\end{figure}

We perform $3$ inference iterations, each using $400$ samples from the updated prior to train the CDF approximator. 
\Cref{fig:lv-atp-posteriors-and-polars} shows the resulting posteriors with default and misspecified support (top row), the discrepancy of $10$ averaged posterior samples from the ground truth (middle row), and the discrepancy of the average summary statistics of the corresponding trajectories, launched from these $10$ samples, from those of the ground-truth trajectory (bottom row). 
We confirm the results of~\Cref{sec:problem-overview,,sec:lv-poc-experiment}: misspecified support adversely affects inference accuracy along the affected parameter dimensions, particularly $\boldsymbol{\theta}_2$, with the effect also propagating to parameters whose support was adequately specified.

\section{Heuristics sensitivity analysis}
\label{appendix:heur-sensitivity}

\begin{figure}[!t]
    \centering
    \includegraphics[width=0.85\columnwidth]{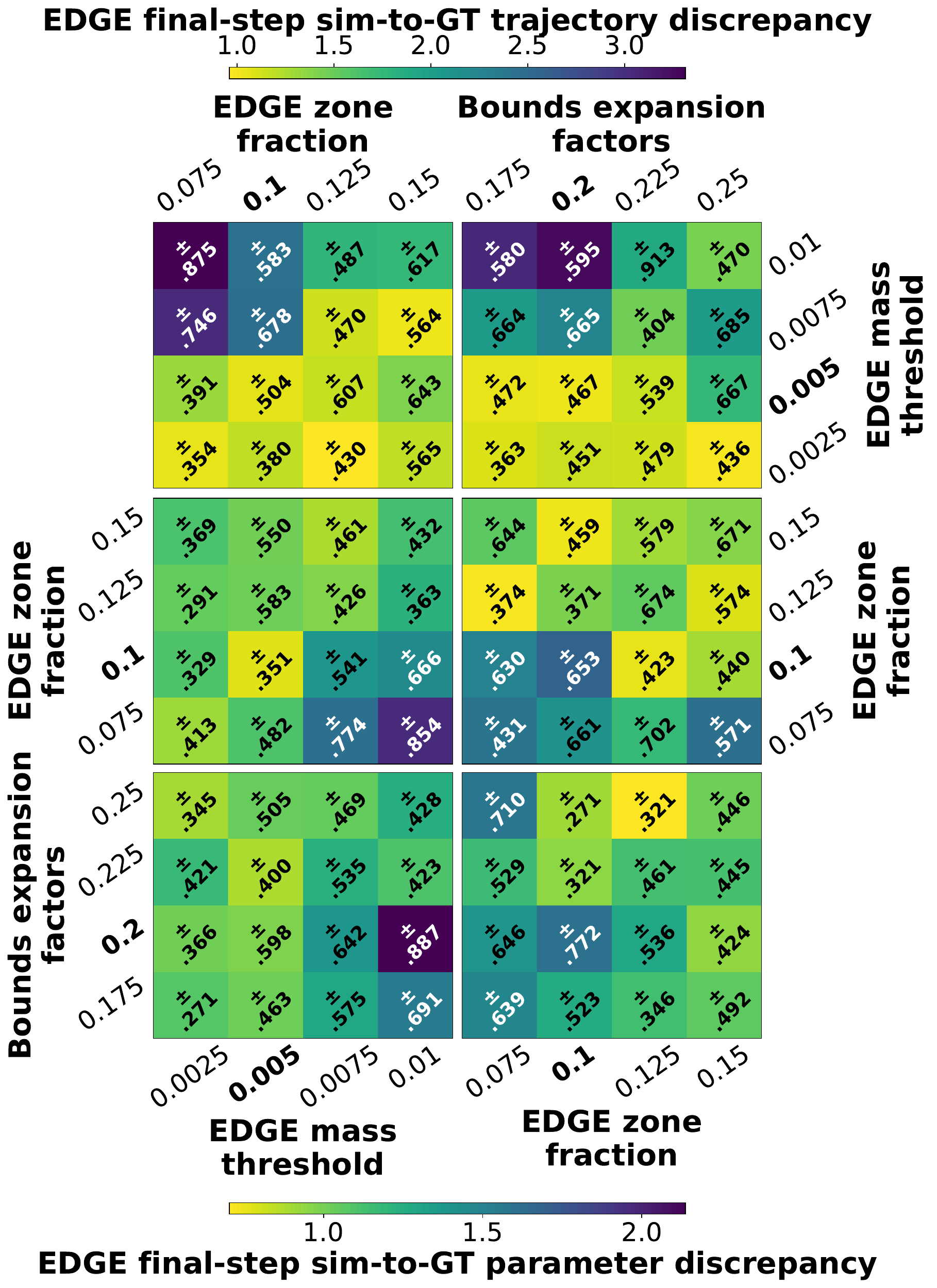}
    \caption{
    EDGE's sensitivity to hyperparameter values, measured as the discrepancy of the final ($\nth{10}$) inference step's result from the ground truth in terms of simulated trajectory and inferred parameter deviations, shown in the upper and lower triangles, respectively. Note the \emph{reversed} colour-coding, as smaller values are better.
    }
    \label{fig:edge-heur-sensitivity-sim2sim-gap}
\end{figure}

\begin{figure}[!t]
    \centering
    \includegraphics[width=1.0\columnwidth]{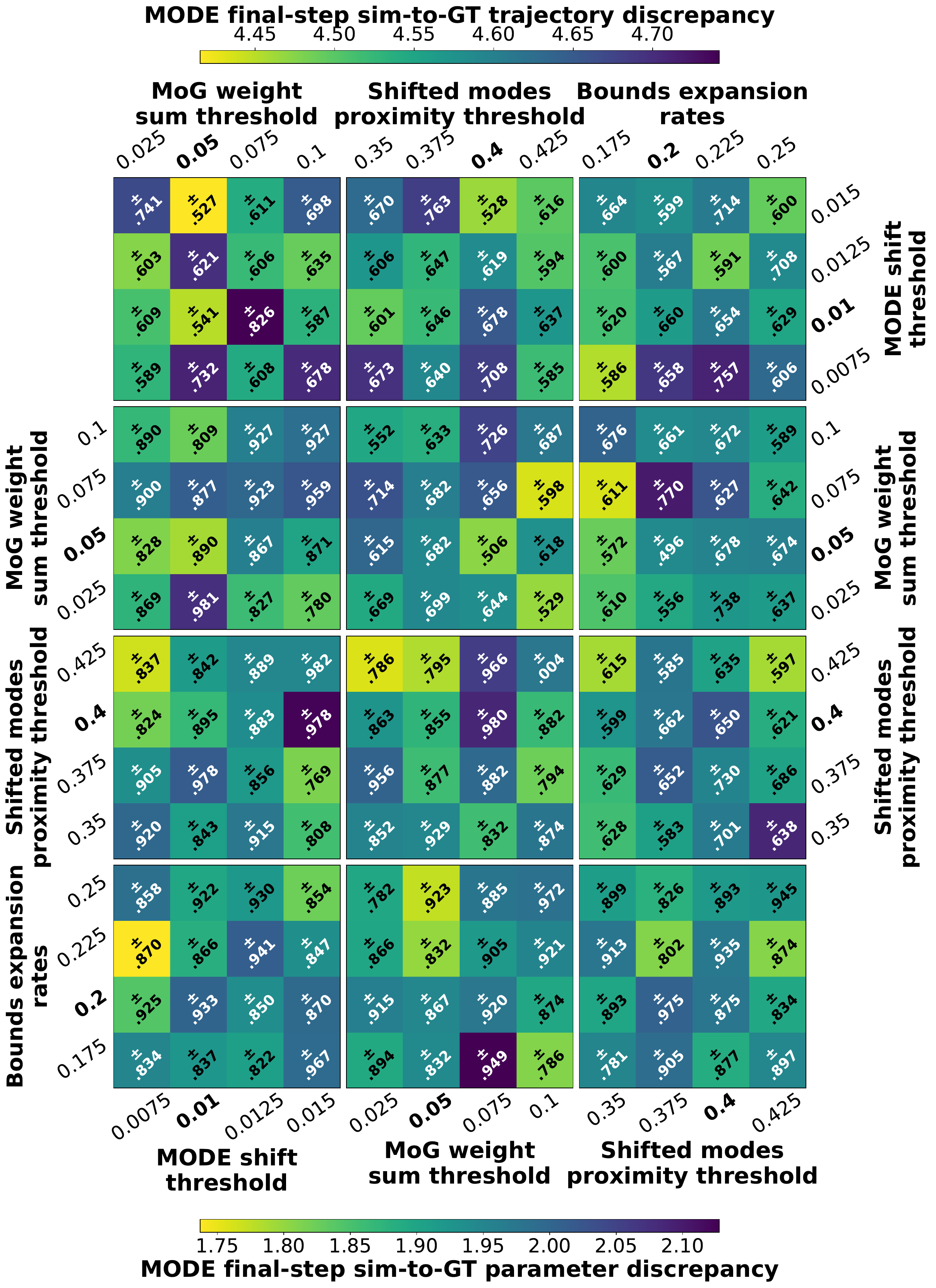}
    \caption{
    MODE hyperparameter sensitivity; upper and lower triangles show the simulated trajectory and the inferred parameter deviations from the ground truth, respectively, for the final ($\nth{10}$) inference step, as in \cref{fig:edge-heur-sensitivity-sim2sim-gap}.
    }
    \label{fig:mode-heur-sensitivity-sim2sim-gap}
\end{figure}

\begin{figure}[!t]
    \centering
    \includegraphics[width=0.8\columnwidth]{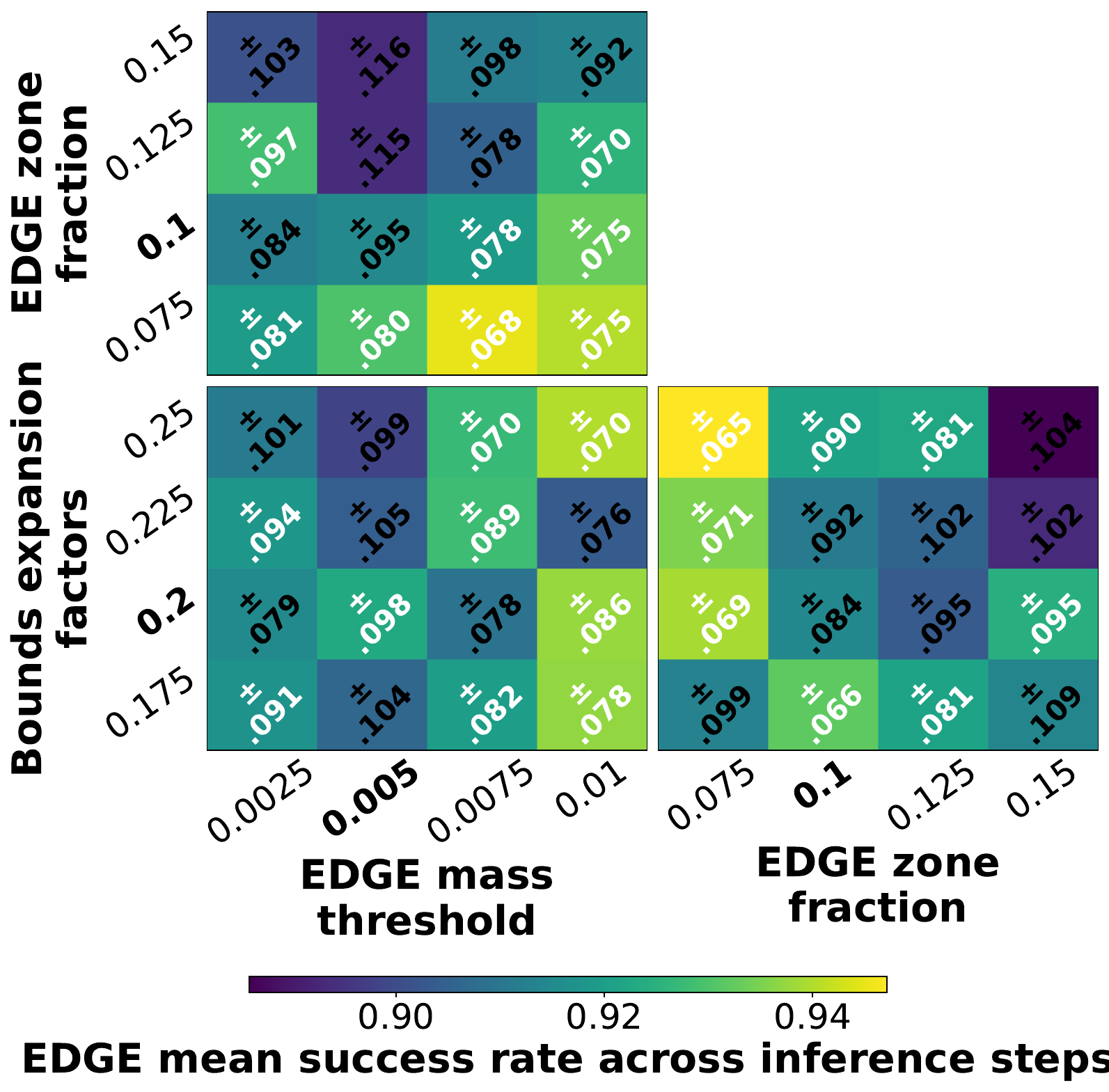}
    \caption{
    EDGE's sensitivity to hyperparameters values measured as the mean simulation success rate (\emph{feasibility}) across inference steps.
    }
    \label{fig:edge-heur-sensitivity-sim-succ-rate}
\end{figure}

\begin{figure}[!t]
    \centering
    \includegraphics[width=1.0\columnwidth]{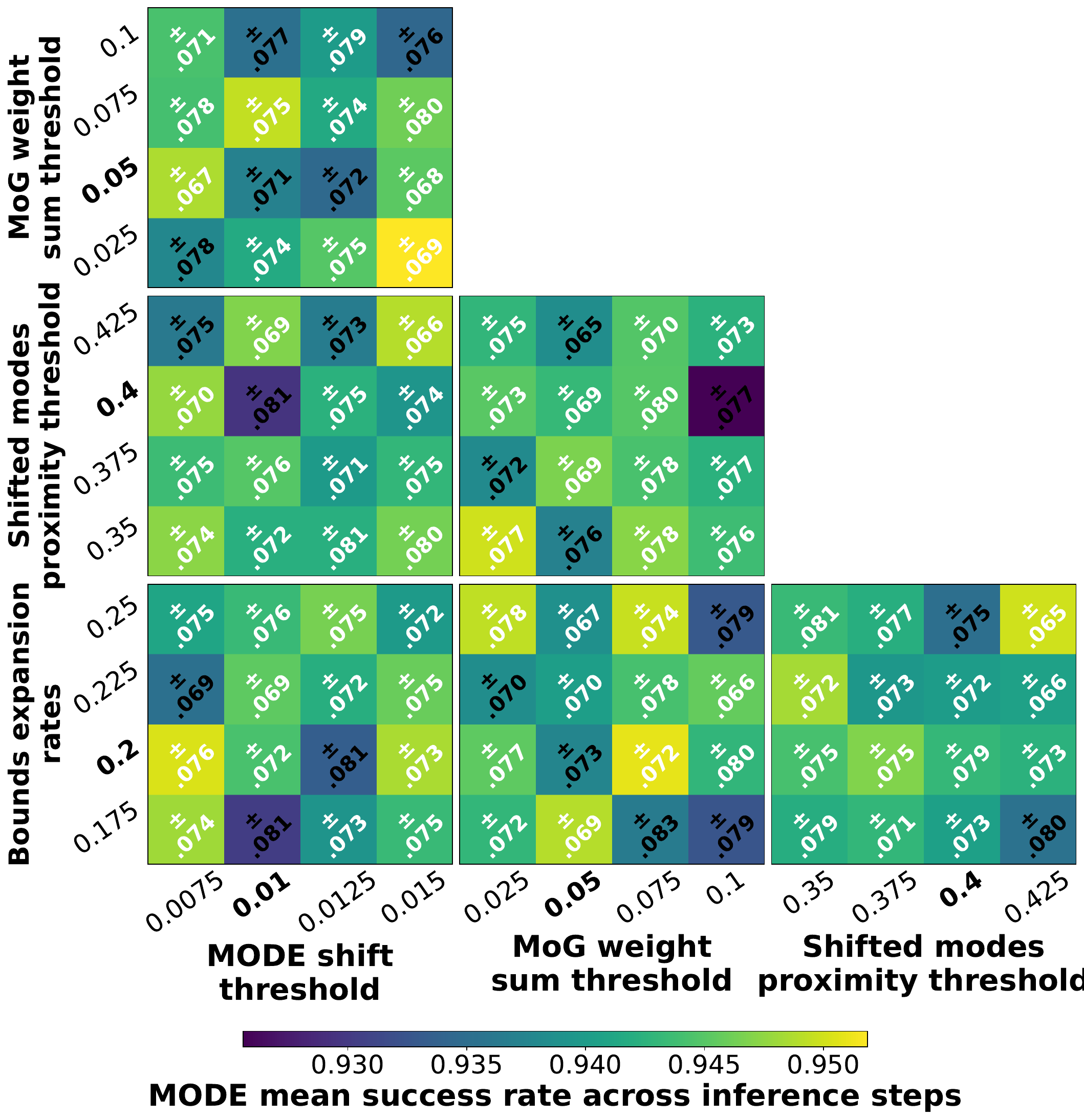}
    \caption{
    MODE's sensitivity to hyperparameters values measured as the mean simulation success rate (\emph{feasibility}) across inference steps.
    }
    \label{fig:mode-heur-sensitivity-sim-succ-rate}
\end{figure}

\Cref{fig:edge-heur-sensitivity-sim2sim-gap,,fig:mode-heur-sensitivity-sim2sim-gap} examine the sensitivity of the EDGE and MODE hyperparameters used in \Cref{sec:lv-poc-experiment}, measured as the discrepancy of the final ($\nth{10}$) inference step's result from the ground truth. 
Grid cells' $\pm$ annotate the standard deviation. 
Similarly, \Cref{fig:edge-heur-sensitivity-sim-succ-rate,,fig:mode-heur-sensitivity-sim-succ-rate} examine the sensitivity measured as the mean simulation success rate (\emph{feasibility}) across inference steps, extending considerations raised in \cref{subsec:widest-prior} and \cref{fig:wide-distr-problm-illustr}. We see that both EDGE and MODE infer more accurately and launch a higher percentage of feasible simulations for the chosen hyperparameter combinations (in bold ticks). In particular, regarding simulation success rate, we see that although hyperparameter choices have a relative impact, the overall success rate is generally $>90\%$. 

\section{Complementary study using the M/G/1 model}
\label{appendix:mg1-case-study}

\subsection{M/G/1 queue model}
\label{subsec:mg1-exper}

The M/G/1 model is a single-server queue with Poisson job arrivals, uniformly distributed service times, $s_i \sim \mathit{U}(\theta_1, \theta_2)$, and inter-arrival times $u_i - u_{i-1} \sim \exp(\theta_3)$.
The server observes only inter-departure times, $d_i - d_{i-1} = s_i +\max(0, u_i - d_{i-1})$, where $u_i$ and $d_i$ are the arrival and departure times of job $i$. Thus, the simulator parameters are $\boldsymbol{\theta} = \{ \theta_1, \theta_2, \theta_3 \}$.

\begin{figure}[t]
    \centering
    \includegraphics[width=1.0\columnwidth]{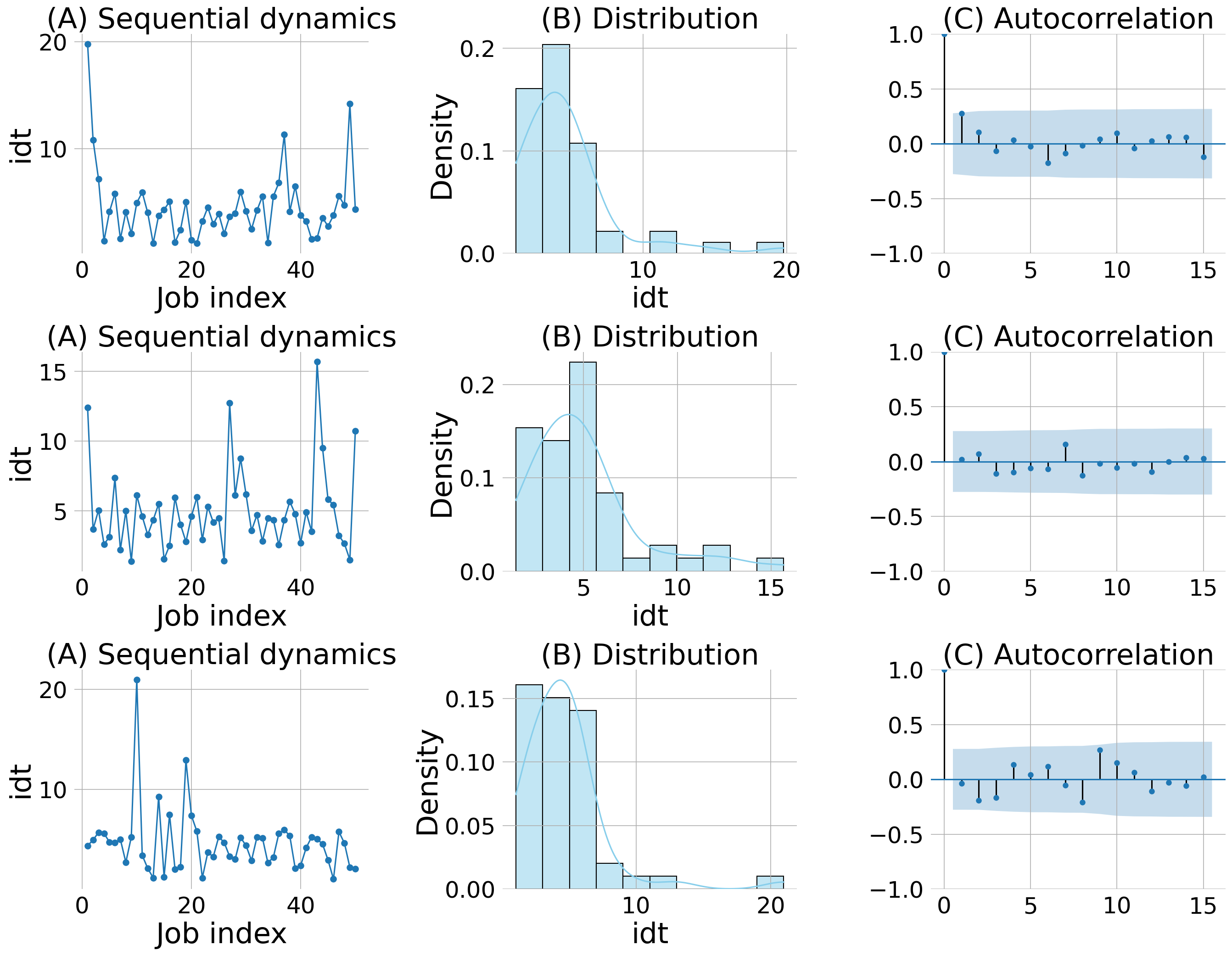}
    \caption{
    Interdeparture time (idt) analysis for $3$ sample M/G/1 system simulations for $50$ jobs, demonstrating stochastic variability in jobs' completion.
    }
    \label{fig:sample-mg1-gt-traj}
\end{figure}

The M/G/1 queue is a stochastic discrete event process and, like Lotka-Volterra, a popular SBI benchmark. Its stochasticity arises from random arrival times and service durations, making the observed data $\mathbf{x}$ less directly informative about the parameters $\boldsymbol{\theta}$, increasing the inference difficulty (\cref{fig:sample-mg1-gt-traj}). For $I=50$ jobs, we observe only $5$ equally spaced percentiles (0th, 25th, 50th, 75th, and 100th) of the empirical distribution of interdeparture times (idts), $d_i - d_{i-1}, i\in[1, I]$, following~\cite{papamakarios2016fast}.

We assume a ground truth of $(\theta_1, \theta_2, \theta_3) = (1.0, 5.0, 0.2)$, 
for which \plotrefblue{$[[0, 10], [0, 10], [0, 0.35]]$} has been shown to be an \plotrefblue{OK} support~\cite{papamakarios2016fast}. 
Given this, a \plotrefred{misspecified} support for $\theta_1$ and $\theta_2$ would be \plotrefred{$[[3.0, 10.0], [0.0, 7.0], [0, 0.35]]$}. We also consider a \plotrefpurple{broad} support being \plotrefpurple{$[[0, 20], [0, 20], [0, 0.5]]$}. The underline colours correspond to~\cref{fig:mg1-misspec-supp-problm-illustr}.

\subsection{The misspecified support issue for M/G/1}
\label{subsec:mg1-misspec-supp-demo}

\begin{figure}[t]
    \centering
    \includegraphics[width=1.0\columnwidth]{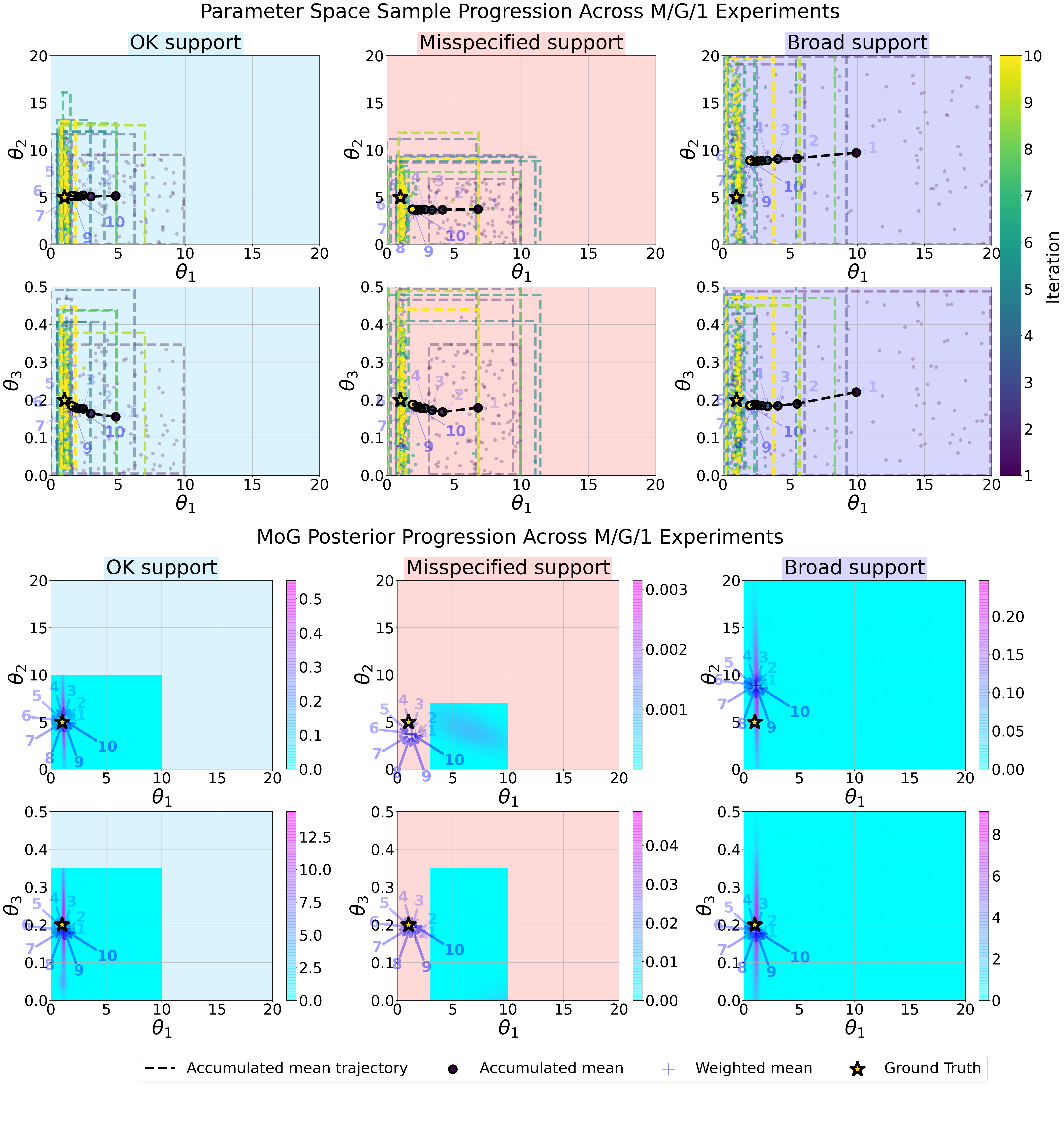}
    \caption{
    The misspecified support issue for M/G/1. 
    We plot the progression of prior samples (top) and their respective MoG posteriors (bottom) along $10$ inference iterations.
    Layout, colour coding, and annotations follow \Cref{fig:misspec-supp-problm-illustr}.
    }
    \label{fig:mg1-misspec-supp-problm-illustr}
\end{figure}

\Cref{fig:mg1-misspec-supp-problm-illustr} shows how a broader support (\plotrefpurple{3rd} col.) hinders the inference of a precise posterior for the M/G/1 task.
We empirically find that M/G/1 is less helpful than Lotka-Volterra in exposing the misspecified support issue, as its formulation is inherently forgiving of outlier samples. The misspecification and excessive broadening implications here are mostly evident in the overall posterior confidence (see heatmap values on the colourbars) and consequently in posteriors' sharpness.

\subsection{Support adaptation for M/G/1 queue}
\label{subsec:mg1-supp-adapt}

\begin{figure}[t]
    \centering
    \includegraphics[width=1.0\columnwidth]{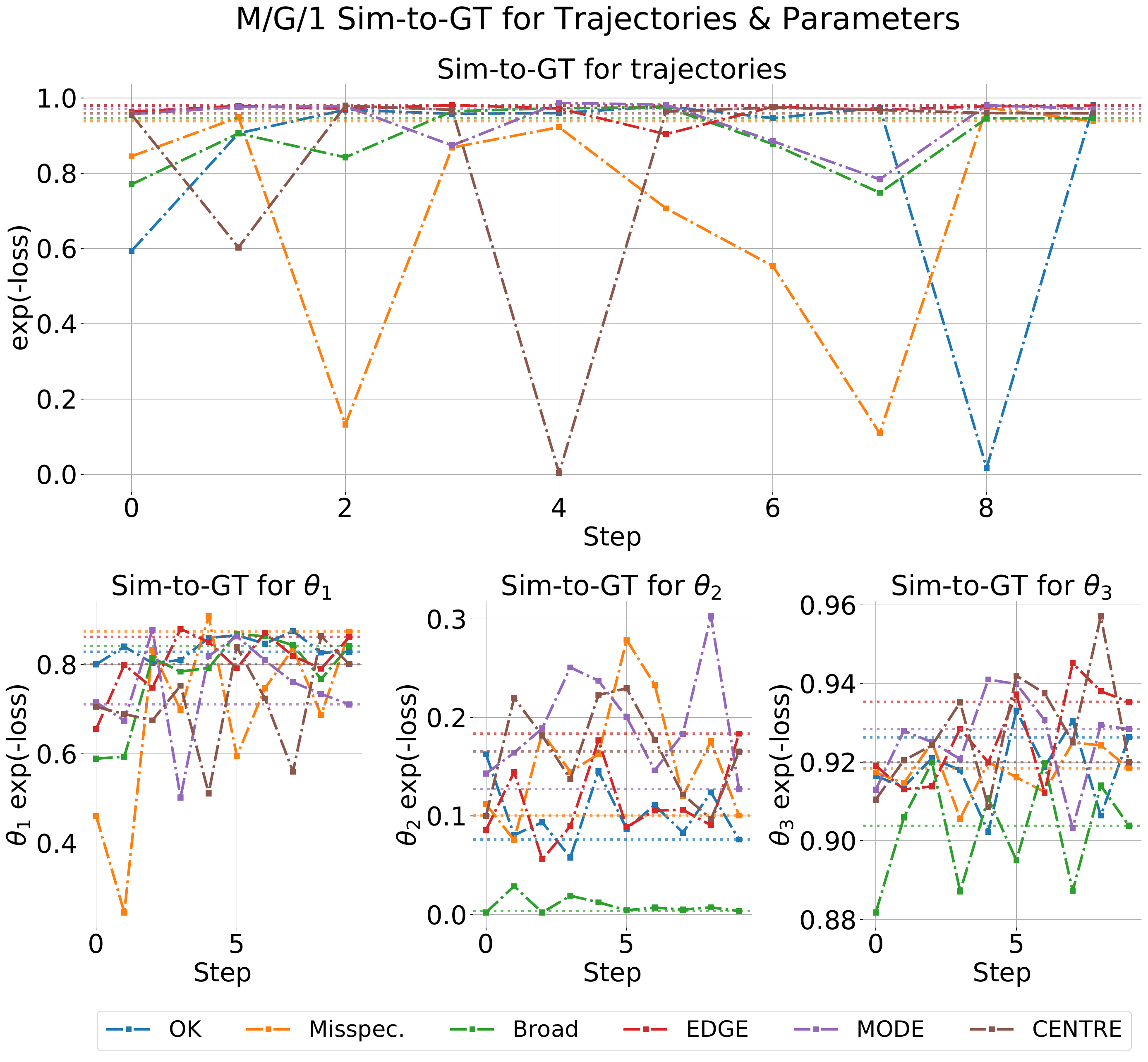}
    \caption{
    M/G/1 Sim2Sim inference exponential of negative loss for trajectories (top) and parameters $\boldsymbol{\theta}$ (bottom grid) over $15$ iterations.
    }
    \label{fig:mg1-res-curves}
\end{figure}

\Cref{fig:mg1-res-curves} corresponds to~\cref{fig:lv-res-curves} discussed in the Lotka-Volterra proof-of-concept (\cref{sec:lv-poc-experiment}), featuring the same inference configurations, distance metrics and $\exp(-\alpha \mathcal{L})$ transformation, and number of posterior samples. The only difference comes to the probability mass threshold ($\tau$) configuration of EDGE, which is now $\tau = 0.001$. 

This time, for M/G/1, it is qualitatively evident that its increased stochasticity poses a bigger challenge for LFI. Although MODE performs better overall in per-parameter inference, EDGE is more consistent among iterations, which is evident in the proximity of the calibrated trajectory to ground truth. 
We also see the adversarial effect of a broad support mainly for $\theta_1$ and $\theta_2$ inference. $\theta_3$ seems to be inferred more robustly overall, possibly because in the chosen domain $\theta_3$ perturbations have a smaller impact on system trajectory. 
We also see that among all the M/G/1 $\boldsymbol{\theta}$ components, the $\theta_1$ inference generally has a stronger qualitative correlation with the overall trajectory approximation.

\bibliographystyle{IEEEtran}
\bibliography{bibliography}

\end{document}